\documentclass[10pt,twocolumn,letterpaper]{article}

\usepackage{cvpr}              

\usepackage[dvipsnames]{xcolor}

\definecolor{cvprblue}{rgb}{0.21,0.49,0.74}
\usepackage[pagebackref,breaklinks,colorlinks,citecolor=cvprblue]{hyperref}
\newcommand{\bx}{\mathbf{x}}

\newcommand{\bmu}{{\boldsymbol{\mu}}}
\newcommand{\bI}{\mathbf{I}}
\newcommand{\bSigma}{{\boldsymbol{\Sigma}}}
\newcommand{\bepsilon}{{\boldsymbol{\epsilon}}}

\newcommand{\vspaceundertab}{\vspace{-.2cm}}

\newcommand{\vspaceundercaption}{\vspace{-0.3cm}}

\usepackage[pagebackref,breaklinks,colorlinks]{hyperref}
\usepackage{arydshln} 
\usepackage[accsupp]{axessibility}

\newcommand{\re}[1]{\textcolor{red}{#1}}

\newcommand{\bl}[1]{\textcolor{blue}{#1}}

\newcommand{\gre}[1]{\textcolor{Green}{#1}}

\def\paperID{5172} 
\def\confName{CVPR}
\def\confYear{2024}

\title{TFMQ-DM: Temporal Feature Maintenance Quantization for Diffusion Models}

\author{Yushi Huang\textsuperscript{1,2}\thanks{Equal Contribution.}~\thanks{Work done during an internship at SenseTime Research.}, Ruihao Gong\textsuperscript{1,2}\footnote[1]{}, Jing Liu\textsuperscript{2,3}, Tianlong Chen\textsuperscript{4}, Xianglong Liu\textsuperscript{1}\thanks{Corresponding Author.}\\
{\small \textsuperscript{1}Beihang University \quad \textsuperscript{2}SenseTime Research \quad \textsuperscript{3}Monash University \quad
\textsuperscript{4}UT Austin}\\
\footnotesize{\texttt{\{huangyushi,gongruihao,liujing5\}@sensetime.com,tianlong.chen@utexas.edu,xlliu@buaa.edu.cn}
}
}
\usepackage{xcolor}
\usepackage{colortbl,booktabs}
\begin{document}
\maketitle
\begin{abstract}
The Diffusion model, a prevalent framework for image generation, encounters significant challenges in terms of broad applicability due to its extended inference times and substantial memory requirements. Efficient Post-training Quantization (PTQ) is pivotal for addressing these issues in traditional models. Different from traditional models, diffusion models heavily depend on the time-step $t$ to achieve satisfactory multi-round denoising. Usually, $t$ from the finite set $\{1, \ldots, T\}$ is encoded to a temporal feature by a few modules totally irrespective of the sampling data. However, existing PTQ methods do not optimize these modules separately. They adopt inappropriate reconstruction targets and complex calibration methods, resulting in a severe disturbance of the temporal feature and denoising trajectory, as well as a low compression efficiency. To solve these, we propose a Temporal Feature Maintenance Quantization (TFMQ) framework building upon a Temporal Information Block which is just related to the time-step $t$ and unrelated to the sampling data. Powered by the pioneering block design, we devise temporal information aware reconstruction (TIAR) and finite set calibration (FSC) to align the full-precision temporal features in a limited time. Equipped with the framework, we can maintain the most temporal information and ensure the end-to-end generation quality. Extensive experiments on various datasets and diffusion models prove our state-of-the-art results. Remarkably, our quantization approach, for the first time, achieves model performance nearly on par with the full-precision model under 4-bit weight quantization. Additionally, our method incurs almost no extra computational cost and accelerates quantization time by $2.0 \times$ on LSUN-Bedrooms $256 \times 256$ compared to previous works. Our code is publicly available at \href{https://github.com/ModelTC/TFMQ-DM}{https://github.com/ModelTC/TFMQ-DM}.

\end{abstract}    
\section{Introduction}
\label{sec:intro}Generative modeling plays a crucial role in machine learning, particularly in applications like image~\cite{kang2023scaling, songddim, ho2020denoising, rombach2022ldm, ho2022classifierfree}, voice~\cite{shen2018natural, ren2022fastspeech}, and text synthesis~\cite{brown2020language, zhang2022opt}. Diffusion models have showcased impressive capabilities in producing high-quality samples across diverse domains. In comparison to generative adversarial networks (GANs)~\cite{goodfellow2020generative} and variational autoencoders (VAEs)~\cite{kingma2022autoencoding}, diffusion models successfully sidestep issues such as model collapse and posterior collapse, resulting in a more stable training process. However, the substantial computational cost poses a critical bottleneck hampering the widespread adoption of diffusion models. Furthermore, the computational cost for diffusion models can be attributed to two primary factors. First, these models typically require hundreds of denoising steps to generate images, rendering the procedure considerably slower than that of GANs. Prior efforts~\cite{songddim, lu2022dpmsolver, kong2021fastdpm, liu2022pseudo} have addressed this challenge by seeking shorter and more efficient sampling trajectories, thereby reducing the number of denoising steps. Second, the substantial network architecture of diffusion models demands considerable time and memory resources, particularly for foundational models pre-trained on large-scale datasets, e.g., LDM~\cite{rombach2022ldm} and Stable Diffusion. Our work aims to tackle the latter challenge, focusing on the compression of diffusion models.

Quantization is currently the most widely used method~\cite{nagel2020adaround, gong2019dsq, nagel2021white, esser2020lsq, bhalgat2020lsq} for compressing models by mapping high-precision floating-point numbers to low-precision numbers. Among different quantization methods, Post-training quantization (PTQ)~\cite{nagel2020adaround, wei2022qdrop, hubara2020improvingadaquant} incurs lower overhead and is more user-friendly without the need for retraining or fine-tuning. While PTQ on conventional models has undergone extensive study~\cite{nagel2020adaround, li2021brecq, wei2022qdrop, gong2019dsq}, its application to diffusion models has shown huge performance degradation, especially under low-bit settings. For instance, Q-Diffusion~\cite{li2023qdiffusion} exhibits severe accuracy drop on some datasets~\cite{yu2016lsun} under 4-bit quantization. We believe the reason they fail to achieve better results is that they all overlook the sampling data independence and uniqueness of temporal features, which are generated from time-step $t$ through a few modules, used to control the denoising trajectory in diffusion models.  Furthermore, we observe that temporal feature disturbance significantly impacts model performance in the aforementioned methods.

To tackle temporal feature disturbance, we first find that the modules generating temporal features are independent of the sampling data and define the whole modules as the Temporal Information Block. All existing methods do not separately optimize this block during the quantization process, causing temporal features to overfit to limited calibration data. On the other hand, since the maximum time-step for denoising is a finite positive integer, the temporal feature and the activations during its generation form a finite set. The optimal approach is to optimize each element in this set individually. Based on these observations and analyses, we propose a novel quantization reconstruction approach, temporal information aware reconstruction~(TIAR), specifically optimizing for temporal features. It aims to reduce temporal feature loss as the optimization objective while isolating the network's components related to sampled data from the generation of temporal features during calibration. Furthermore, we also introduce a calibration strategy, finite set calibration~(FSC), for the finite set of temporal features and activations during its generation. This strategy employs different quantization parameters for activations corresponding to different time-steps. Moreover, the calibration speed of this method is faster than existing mainstream methods~\cite{esser2020lsq, bhalgat2020lsq}, for example, we speedup quantization time by $2.0\times$ on LSUN-Bedrooms $256\times 256$ dataset, yet the strategy incurs negligible inference and storage overhead. The overview of our framework can be seen in Fig. \ref{overview}. In summary, our contributions are as follows:
\begin{itemize}
    \item {We discover that existing quantization methods suffer from temporal feature disturbance, disrupting the denoising trajectory of diffusion models and significantly affecting the quality of generated images.}
    \item {We reveal that the disturbance comes from two aspects: inappropriate reconstruction target and unaware of finite activations. Both inducements ignore the special characteristics of time information-related modules.}
    \item {An advanced framework~(TFMQ-DM) is proposed, consisting of temporal information aware reconstruction~(TIAR) for weight quantization and finite set calibration~(FSC) for activation quantization. Both are based on a Temporal Information Block specially devised for diffusion models.}
    \item {Extensive experiments on various datasets show that our novel framework achieves a new state-of-the-art result in PTQ of diffusion models, especially under 4-bit weight quantization, and significantly accelerates quantization time. For some hard tasks, e.g., CelebA-HQ $256 \times 256$, our method reduces the FID score by 6.71 (images in appendix).}
\end{itemize}

\section{Related Work}
\label{sec:relatedwork}

\subsection{Efficient Diffusion Models}
There are diverse perspectives to accelerate the inference of diffusion model, e.g., retraining-based model design~\cite{chung2022come, lyu2022accelerating, zheng2022truncated, franzese2022much} and retraining-free sampler strategy~\cite{kong2021fastdpm, songddim, zhang2022gddim}. However, the retraining-based method proves resource-intensive and time-consuming. The efficient samplers can reduce sampling iterations but still suffer from diffusion models' extensive parameters and computational complexity. In this paper, we focus on diminishing the time and memory overhead of the single-step denoising process using low-bit quantization in a training-free manner, a method orthogonal to previous speedup techniques.

\subsection{Model Quantization}

Quantization is a predominant technique for minimizing storage and computational costs. It can be categorized into quantization-aware training (QAT)~\cite{gong2019dsq, jacob2018quantization, louizos2018relaxed, zhang2023root, zhuang2018towards} and post-training quantization (PTQ)~\cite{hubara2020improvingadaquant, li2021brecq, lin2022fqvit, nagel2020adaround, wei2022qdrop}. QAT requires intensive model training with substantial data and computational demands. Correspondingly, PTQ compresses models without re-training, making it a preferred method due to its minimal data requirements and easy deployment on real hardware. In PTQ, high-precision values are mapped into discrete levels using uniform quantization expressed as:
\begin{equation}
    \hat{x} = \Phi ( \lfloor \frac{x}{s} \rceil + z, 0, 2^{b}-1),
    \label{eq:uniform_quant}
\end{equation}
where $x$ represents floating-point data, $\hat{x}$ is the quantized value, $s$ is the quantization step size, $z$ is the zero offset, and $b$ is the target bit-width. The clamp function $\Phi(\cdot)$ clips the rounded value $\left\lfloor \frac{x}{s} \right\rceil + z$ within the range of $[0, 2^{b}-1]$. However, naive quantization may lead to accuracy degradation, especially for low-bit quantization. Recent studies~\cite{nagel2020adaround,hubara2020improvingadaquant, wang2020towards,li2021brecq,wei2022qdrop} have explored innovative strategies based on reconstruction to preserve model performance after low-bit quantization. 

In contrast, the iterative denoising process in diffusion models presents new challenges for PTQ in comparison to traditional models. PTQ4DM~\cite{shang2022ptq4dm} represents the initial attempt to quantize diffusion models to 8-bit, albeit with limited experiments and lower resolutions. Conversely, Q-Diffusion~\cite{li2023qdiffusion} achieves enhanced performance and is evaluated on a broader dataset range. Moreover, PTQD~\cite{he2023ptqd} eliminates quantization noise through correlated and residual noise correction. Notably, traditional single-time-step PTQ calibration methods are unsuitable for diffusion models due to significant activation distribution changes with each time-step~\cite{shang2022ptq4dm, li2023qdiffusion, wang2023towards, so2023temporal}. ADP-DM~\cite{wang2023towards} proposes group-wise quantization across time-steps for diffusion models, and TDQ~\cite{so2023temporal} introduces distinct quantization parameters for different time-steps. However, all of the above works overlook the specificity of temporal features. To address temporal feature disturbance in the aforementioned works, our study delves into the inducements of the phenomenon and introduces a novel reconstruction and calibration framework, significantly enhancing quantized diffusion model performance.
\section{Preliminaries}\label{pre}
\begin{figure*}[t]
    \centering
    \setlength{\abovecaptionskip}{0.2cm}
    \includegraphics[width=\textwidth]{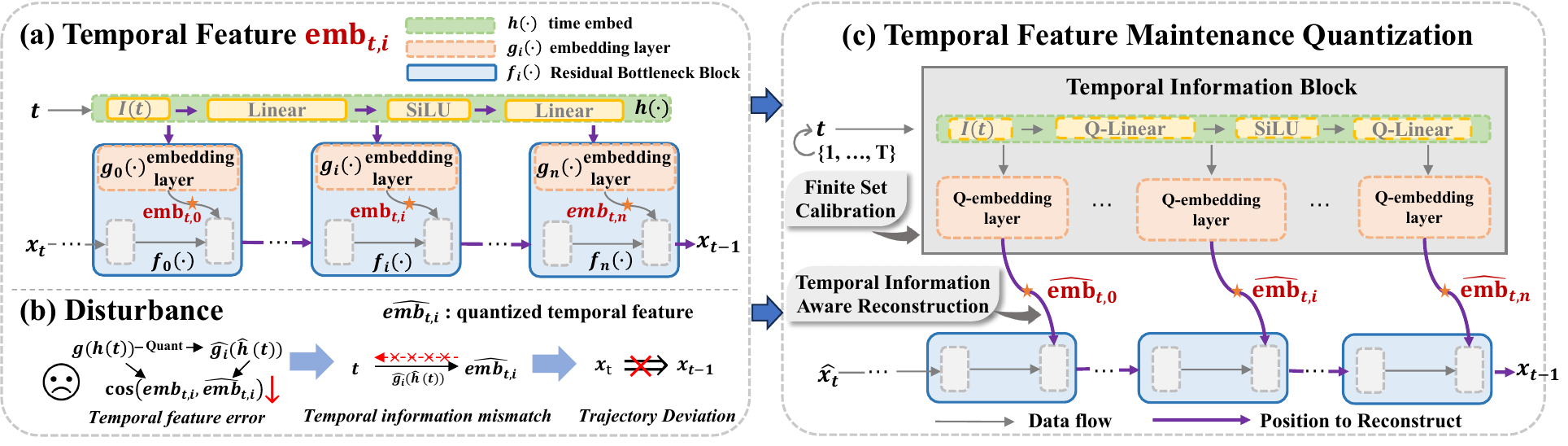}
     \caption{Overview of the proposed Temporal Feature Maintenance Quantization. (a) Temporal Feature $\mathbf{emb}_{t,i}$, belonging to a finite set representing temporal information, has been overlooked in previous works due to inappropriate reconstruction targets (box with a solid line). (b) This oversight leads to a severe disturbance for $\mathbf{emb}_{t,i}$ and results in the mismatch of crucial temporal information for the diffusion model's generation, causing a deviation in the denoising trajectory and a significant drop in accuracy. (c) Based on these analyses, we introduce a Temporal Information Block that exclusively correlates with the time-step $t$. Leveraging this $\bx_t$-unrelated block, we enable Temporal Information Aware Reconstruction and Finite Set Calibration (utilizing the finite number of $t$). This approach achieves the maintenance of temporal features and yields state-of-the-art results.}
    \label{overview}
    \vspaceundercaption
\end{figure*}
\noindent\textbf{Diffusion models.} Diffusion models~\cite{songddim, ho2020denoising} iteratively add Gaussian noise with a variance schedule $\beta_1, \ldots, \beta_T \in (0, 1)$ to data $\bx_0 \sim q(\bx)$ for $T$ times as sampling process, resulting in a sequence of noisy samples $\bx_1, \ldots, \bx_T$. In DDPMs~\cite{ho2020denoising}, the former sampling process is a Markov chain, taking the form:
\begin{equation}
    \begin{split}
        q(\bx_{1:T} | \bx_0) & = \prod_{t=1}^T q(\bx_t | \bx_{t-1} ), \\
        q(\bx_t|\bx_{t-1}) & = \mathcal{N}(\bx_t;\sqrt{\alpha_t}\bx_{t-1}, \beta_t \bI),
        \label{eq:sampling}
    \end{split}
\end{equation}
where $\alpha_t=1-\beta_t$. Conversely, the denoising process removes noise from a sample from Gaussian noise $\bx_T \sim \mathcal{N}(\mathbf{0}, \mathbf{I})$ to gradually generate high-fidelity images. Nevertheless, due to the unavailability of the true reverse conditional distribution $q(\bx_{t-1} | \bx_t)$, diffusion models approximate it via variational inference by learning a Gaussian distribution $p_\theta(\bx_{t-1} | \bx_t)=\mathcal{N}(\bx_{t-1}; \bmu_\theta(\bx_t, t), \bSigma_\theta(\bx_t, t))$, the $\bmu_\theta$ can be derived by reparameterization trick as follows:
\begin{equation}
    \bmu_\theta(\bx_t, t)  = \frac{1}{\sqrt{\alpha_t}}\left( \bx_t - \frac{\beta_t}{\sqrt{1-\bar\alpha_t}} \bepsilon_\theta(\bx_t, t) \right), \label{eq:ddpm_reverse_mean}
\end{equation}
where $\bar\alpha_t = \prod_{i=1}^t \alpha_i$ and $\bepsilon_\theta(\cdot)$ is a noise estimation model. The variance $\bSigma_\theta(\bx_t, t)$ can be either learned~\cite{nichol2021improvedDDPM} or fixed to a constant schedule~\cite{ho2020denoising} $\sigma_t$. When employing the latter method, $\bx_{t-1}$ can be expressed as:
\begin{equation}
\bx_{t-1} = \frac{1}{\sqrt{\alpha_t}}\left( \bx_t - \frac{\beta_t}{\sqrt{1-\bar\alpha_t}} {\bepsilon}_\theta(\bx_t, t) \right) + \sigma_t \mathbf{z},
\label{eq:denoising}
\end{equation}
where $\mathbf{z} \sim \mathcal{N}(\mathbf{0}, \mathbf{I})$.

\vspace{1em}
\noindent\textbf{Reconstruction on diffusion models.}\label{recon} UNet~\cite{ronneberger2015unet}, the predominant model employed as $\bepsilon_\theta(\cdot)$ in Eq.~\ref{eq:denoising} to predict Gaussian noise, can be divided into blocks that incorporate residual connections (such as Residual Bottleneck Blocks or Transformer Blocks~\cite{peebles2023scalable}) and the remaining layers. Numerous PTQ approaches for diffusion models are grounded in layer/block-wise reconstruction~\cite{shang2022ptq4dm, he2023ptqd, li2023qdiffusion, so2023temporal} to obtain optimal quantization parameters. For example, in the Residual Bottleneck Block, this approach typically minimizes the following loss function as its optimization objective:
\begin{equation}
    \mathcal{L}_i=\| f_i(\cdot) - \widehat{f_i}(\cdot)\|_{F}^2,
    \label{lossfunction}
\end{equation}
where $\|\cdot\|^2_{F}$ denotes the Frobenius norm. The function $f_i(\cdot)$ represents the $i^{\text{th}}$ Residual Bottleneck Block, and $\widehat{f_i}(\cdot)$ is its quantized counterpart. Furthermore, in the ensuing sections, we use $n$ to denote the total number of Residual Bottleneck Blocks in a single diffusion model.

\vspace{1em}
\noindent\textbf{Temporal feature in diffusion models. } Seeing from Fig.~\ref{overview} (a), time-step $t$ is encoded with \verb|time embed|\footnote{\href{https://github.com/CompVis/stable-diffusion/blob/main/ldm/modules/diffusionmodules/openaimodel.py\#L507}{PyTorch time embed implementation in diffusion models.}} and then passes through the \verb|embedding layer|\footnote{\href{https://github.com/CompVis/stable-diffusion/blob/main/ldm/modules/diffusionmodules/openaimodel.py\#L218}{PyTorch embedding layer implementation in diffusion models.}} in each Residual Bottleneck Block, resulting in a series of unique activations. In this paper, we denote these activations as temporal features. Notably, temporal features are independent of $\bx_t$ and unrelated to other temporal features from different time-steps. To enhance clarity, we simplify our notation as follows: we represent \verb|time embed| as $h(\cdot)$, the \verb|embedding layer| in the $i^{\text{th}}$ Residual Bottleneck Block as $g_i(\cdot)$, and denote the $i^{\text{th}}$ temporal feature at time-step $t$ as $\mathbf{emb}_{t,i}$. Moreover, as illustrated in Fig.~\ref{overview} (a), the relationship is explicitly expressed by the equation:
\begin{equation}
\text{Temporal feature}: \mathbf{emb}_{t,i} = g_i(h(t))
\label{emb_rel}
\end{equation}

Additionally, we have found that temporal features play a crucial role in the context of the diffusion model, holding unique and substantial physical implications. These features encompass temporal information that signifies the current image's temporal position along the denoising trajectory. Within the UNet structure, each time-step transforms into these temporal features, thereby controlling the denoising trajectory by applying them to the features of images generated at each iteration.

\section{TFMQ for Diffusion Models}
We present our novel PTQ  framework in this section. 
We first observe temporal disturbance in previous methods in Sec.~\ref{sec_dist} and then analyze the inducements in Sec.~\ref{inducement}. Finally, we propose our solutions in Sec.~\ref{maintenance}.
\subsection{Temporal Feature Disturbance}\label{sec_dist}
Based on Sec.~\ref{pre}, we investigate the impact of previous PTQ works on temporal features, and we identify the phenomenon of temporal feature disturbance, which significantly deteriorates the quality of generated images. 

\begin{figure}[tp!]
   \centering
    \setlength{\abovecaptionskip}{0.2cm}
    \begin{subfigure}[tp!]{0.236\textwidth}
        \centering
        \includegraphics[width=\textwidth]{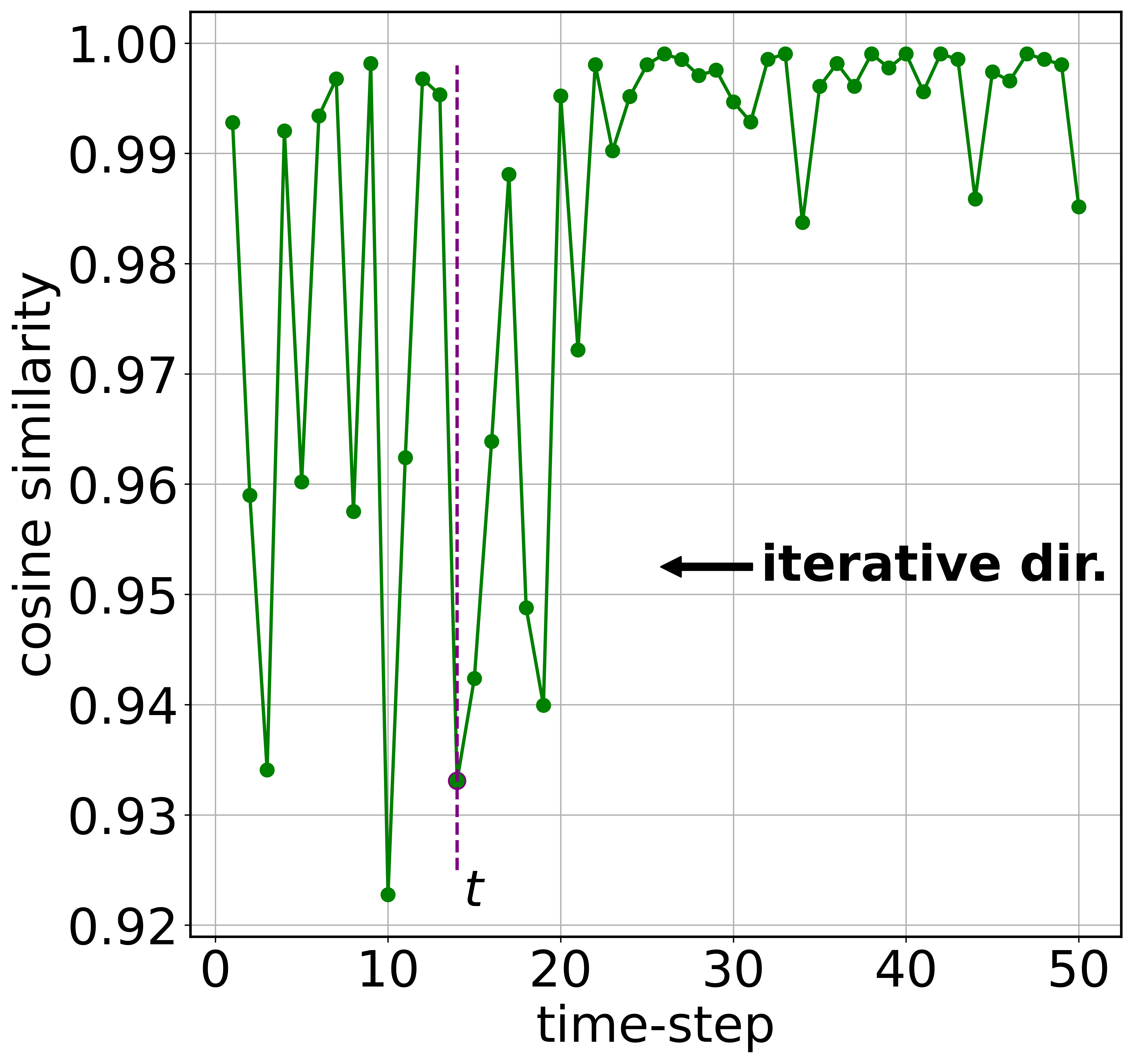}
    \end{subfigure}
    \begin{subfigure}[tp!]{0.236\textwidth}
        \centering
        \includegraphics[width=\textwidth]{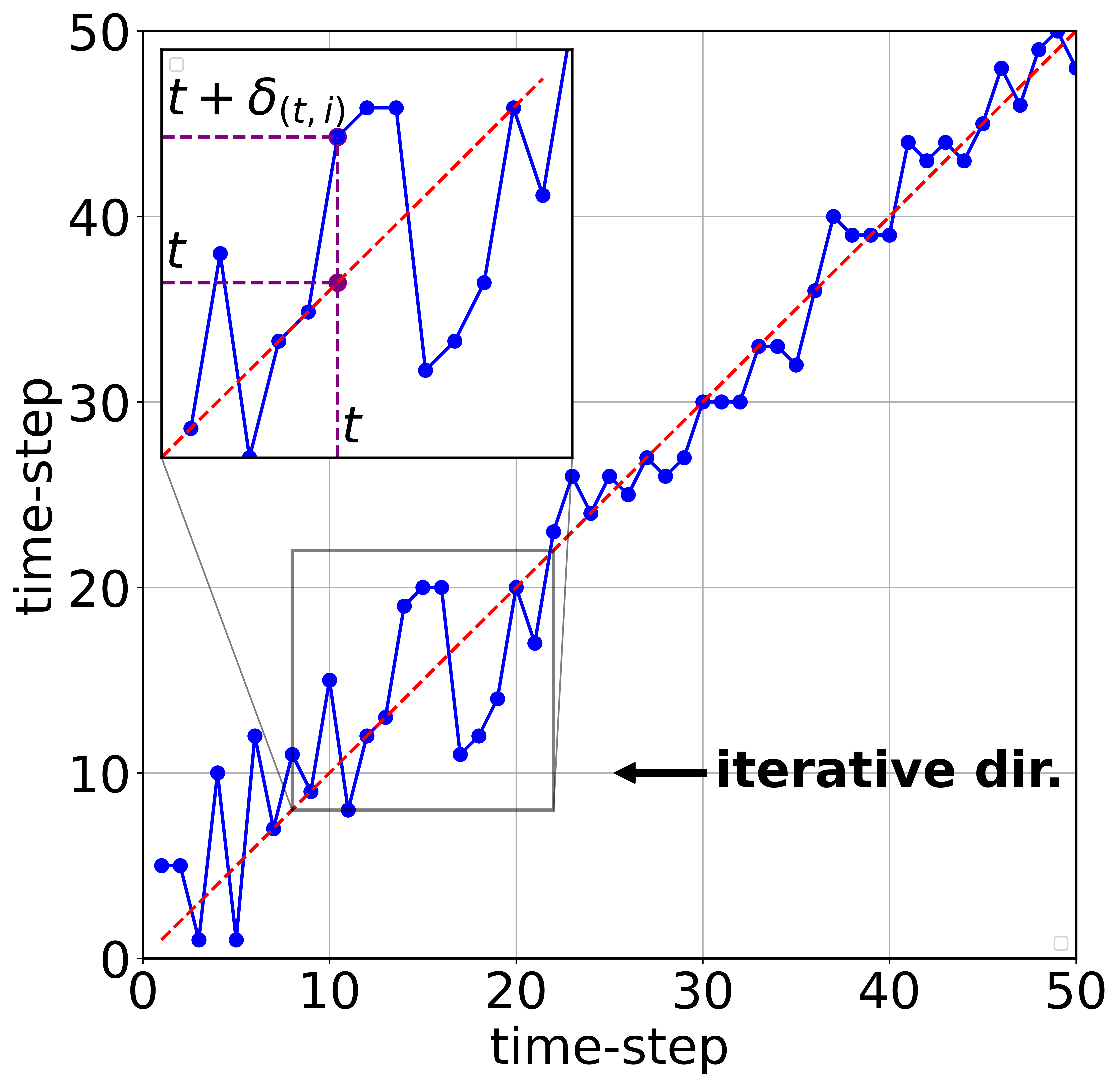}
    \end{subfigure}
    \caption{(Left)~Temporal feature disturbance. The inflection points serve as indicators of temporal feature errors at different time-steps, and they highlight the significant phenomenon of temporal feature disturbance. (Right)~Temporal information mismatch. The coordinates of the inflection points on the blue curve can denoted as $(t, t+\delta_{t, i})$. It indicates $\mathbf{emb}_{t+\delta_{t, i}, i}$ exhibits the highest similarity with $\widehat{\mathbf{emb}_{t,i}}$.}
    \label{disturb-mismatch}
    \vspaceundercaption
\end{figure}

\vspace{1em}
\noindent\textbf{Temporal feature error.} We thoroughly analyze temporal feature variations before and after the quantization of \verb|embedding layers| and \verb|time embed| in the Stable Diffusion model $(T=50, i = 11)$. Prior to this analysis, we introduce the temporal feature error as defined by:
\begin{equation}
\text{cos}(\mathbf{emb}_{t,i}, \widehat{\mathbf{emb}_{t,i}}),
\label{temporal_feature_error}
\end{equation}
where $\text{cos}(\cdot)$ denotes cosine similarity, and $\widehat{\mathbf{emb}_{t,i}}$ signifies the temporal feature corresponding to $\mathbf{emb}_{t,i}$ in the quantized model. As illustrated in Fig.~\ref{disturb-mismatch}~(Left), quantization induces notable temporal feature errors. We term this phenomenon, characterized by substantial temporal feature errors within diffusion models, as temporal feature disturbance.

\vspace{1em}
\noindent\textbf{Temporal information mismatch.} Temporal feature disturbance alters the original embedded temporal information. Specifically, $\mathbf{emb}_{t, i}$ is intended to correspond to time-step $t$. However, due to significant errors, the quantized model's $\widehat{\mathbf{emb}_{t, i}}$ is no longer accurately associated with $t$, resulting in what we term as temporal information mismatch: 
\begin{equation}
    t \leftarrow \mathbf{emb}_{t,i},\quad t \nleftarrow \widehat{\mathbf{emb}_{t,i}}. \\
    \label{misma}
\end{equation}
Furthermore, as depicted in Fig.~\ref{disturb-mismatch}~(Right), we even observe a pronounced temporal information mismatch. Specifically, the temporal feature generated by the quantized model at time-step $t$ exhibits a divergence from that of the full-precision model at the corresponding time-step, Instead, it tends to align more closely with the temporal feature corresponding to $t + \delta_t$, importing wrong temporal information from $t+\delta_t$.

\begin{figure*}[t]
    \centering
    \setlength{\abovecaptionskip}{0.2cm}
     \includegraphics[width=\textwidth]{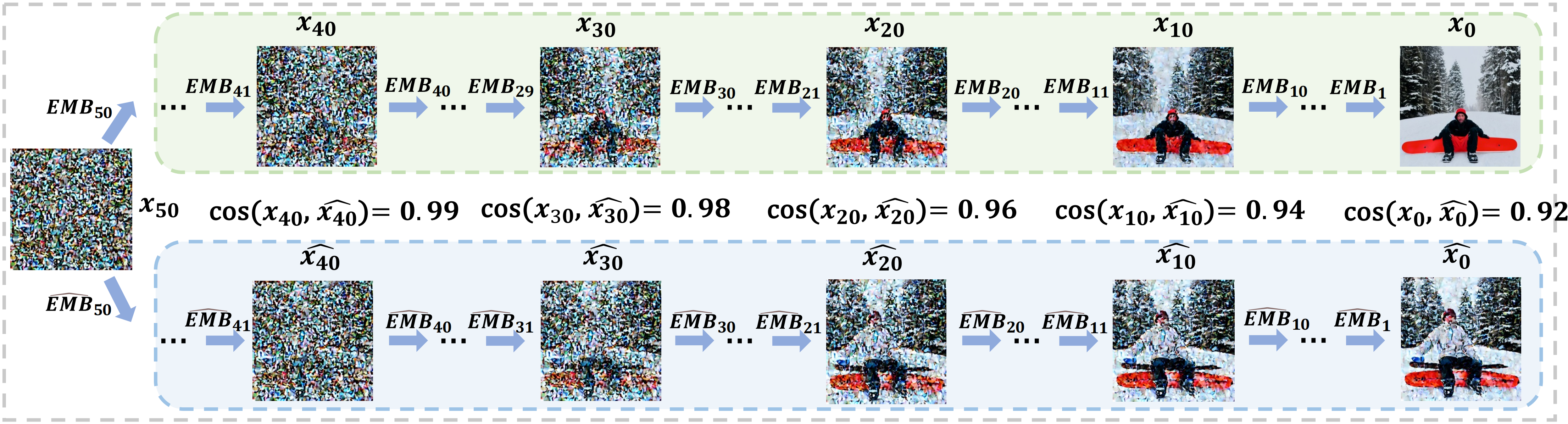}
     \caption{Denoising process of full-precision~(Upper) and w4a8 quantized~(Lower) Stable-Diffusion $(T = 50)$ under the same experiment settings and prompt: \textit{A man in the snow on a snow board}. We represent $\{\mathbf{emb}_{t,i}\}_{i = 0, \ldots, n}$ and $\{\widehat{\mathbf{emb}_{t,i}}\}_{i = 0, \ldots, n}$ as $\mathbf{EMB}_t$ and $\widehat{\mathbf{EMB}_t}$, respectively. Additionally, we denote $\widehat{\bx_t}$ as $\bx_t$ in the context of the quantized diffusion model. It is noteworthy that, in the quantized model employed here, to showcase the impact of temporal features, only the layers in Temporal Information Block are quantized and the components unrelated to the generation of temporal features are maintained in full precision.}
    \label{denoising_compare}
    \vspaceundercaption
\end{figure*}

\vspace{1em}
\noindent\textbf{Trajectory deviation.} Temporal information mismatch delivers wrong temporal information, therefore, causing a deviation in the corresponding temporal position of the image within the denoising trajectory, ultimately leading to:
\begin{equation}
    \bx_t \nRightarrow \bx_{t-1},
    \label{devi}
\end{equation}
where we apply disrupted temporal features to the model. Evidently, as the deviation in the denoising trajectory accumulates with the increase in the number of denoising iterations, the final generated image struggles to align with $\bx_0$. This evolution is illustrated in Fig.~\ref{denoising_compare}, where we maintain UNet excluding \verb|embedding layers| and \verb|time embed| in full precision.
\subsection{Inducement Analyses}\label{inducement}
In this section, we explore the two inducements of temporal feature disturbance. For the purpose of clarity, in the subsequent sections, ``reconstruction" specifically points to slight weight adjustment for minimal quantization error, while ``calibration" specifically refers to activation calibration.

\vspace{1em}
\noindent\textbf{Inappropriate reconstruction target.} Previous PTQ works~\cite{he2023ptqd, li2023qdiffusion, so2023temporal} have achieved remarkable progress on diffusion models. However, these existing methods overlook the temporal feature's independence and its distinctive physical significance. During their reconstruction processes, there was a lack of optimization for the \verb|embedding layers|; instead, a Residual Bottleneck Block of coarser granularity was selected as the reconstruction target. This method involves two potential factors causing temporal feature disturbance:
\begin{itemize}
    \item {Optimize the objective as expressed in Eq.~\ref{lossfunction} to decrease the reconstruction loss of Residual Bottleneck Block, as opposed to direct reduction of temporal feature loss.}
    \item {During backpropagation of the reconstruction process, \verb|embedding layers| independent from $\bx_t$ are affected by $\bx_t$, resulting in an overfitting scenario on limited calibration data.}
\end{itemize}

To further validate our analyses, we respectively evaluate the FID~\cite{heusel2018gans} and sFID~\cite{salimans2016improved} for this reconstruction method, \emph{e.g.}, BRECQ~\cite{li2021brecq} and the approach where we freeze the parameters of the \verb|embedding layers| during the reconstruction of the Residual Bottleneck Block, initializing the \verb|embedding layers| solely through Min-max~\cite{nagel2021white} for comparison. As shown in Tab.~\ref{tab:recon_comp}, the Freeze strategy exhibits better results, which verify that \verb|embedding layers| serve as their own optimization objective and maintain their independence of $\bx_t$ can significance mitigate temporal feature disturbance, especially at low-bit.
\begin{table}[!h]\setlength{\tabcolsep}{10pt}
  \centering
  \caption{FID and sFID on LSUN-Bedrooms $256\times256$~\cite{yu2016lsun} for LDM-4 with 50000 sampling images. Prev represents BRECQ. Freeze denotes our trial.} 
  \resizebox{0.9\linewidth}{!}{
  \begin{tabular}{lcll}
    \toprule
    \textbf{Methods} & \textbf{Bits (W/A)} & FID$\downarrow$ & sFID$\downarrow$ \\
    \midrule
    Full Prec. & 32/32 & 2.98 & 7.09 \\
    \midrule
    Prev & 8/8 & 7.51 &  12.54 \\
    \rowcolor[gray]{0.9}Freeze & 8/8 & \textbf{5.76~\bl{(-1.75)}} & \textbf{8.42~\bl{(-4.12)}} \\
    \midrule
    Prev & 4/8  & 9.36 & 22.73 \\
    \rowcolor[gray]{0.9}Freeze & 4/8 & \textbf{7.08~\bl{(-2.28)}} & \textbf{16.82~\bl{(-5.91)}} \\
    \bottomrule
\end{tabular}}
    \label{tab:recon_comp}
\end{table}

\vspace{1em}
\noindent\textbf{Unaware of finite activations within $h(\cdot)$ and $g_i(\cdot)$.} We observe that, given $T$ as a finite positive integer, the set of all possible activation values for \verb|embedding layers| and \verb|time embed| is finite and strictly dependent on time-steps. Within this set, activations corresponding to the same layer display notable range variations across different time-steps (refer to the appendix). Previous methods~\cite{so2023temporal, wang2023towards} mainly focus on finding the optimal calibration method for $\widehat\bx_t$-related network components. Moreover, akin to the first inducement, their calibration is directly towards the Residual Bottleneck Block, which proves suboptimal (refer to the appendix). However, based on the finite activations, we can employ calibration methods, especially for these time information-related activations, to better adapt to their range variations. 

\subsection{Temporal Feature Maintenance}\label{maintenance}
To address the problem of temporal feature disturbance, we design a novel Temporal Information Block to maintain the temporal features. Based on the block, Temporal Information Aware Reconstruction and Finite Set Calibration are proposed to solve the two inducements analyzed above.

\vspace{1em}
\noindent\textbf{Temporal Information Block.} Based on the inducements, it is crucial to meticulously separate the reconstruction and calibration process for each \verb|embedding layer| and Residual Bottleneck Block to enhance quantized model performance. Considering the unique structure of the UNet, we consolidate all \verb|embedding layers| and \verb|time embed| into a unified Temporal Information Block, which can be denoted as $\{g_i(h(\cdot))\}_{i=0, \ldots, n}$ (see Fig. \ref{overview} (c)).

\vspace{1em}
\noindent\textbf{Temporal information aware reconstruction.} Based on the Temporal Information Block, we propose Temporal information aware reconstruction (TIAR) to tackle the first inducement. The optimization objective for the block during the reconstruction process is as follows:
\begin{equation}
     \mathcal{L}_{TIAR}=\sum_{i=0}^{n}\| g_i(h(t)) -  \widehat{g}_i(\widehat{h}(t)) \|_{F}^2,
    \label{time_embedding_block_lossfunction}
\end{equation}
where $\widehat{h}(\cdot)$ and $\widehat{g}_i(\cdot)$ are quantized versions of $h(\cdot)$ and $g(\cdot)$, respectively. With this reconstruction, weights are adjusted to pursue a minimal disturbance for temporal features.


\vspace{1em}
\noindent\textbf{Finite set calibration.} To address the challenge posed by the wide span of activations within a finite set for the second inducement, we propose Finite Set Calibration (FSC) for activation quantization. This strategy employs $T$ sets of quantization parameters for every activation within all the \verb|embedding layers| and \verb|time embed|, such as $\{(s_T, z_T), \ldots, (s_1, z_1)\}$ for activation $\bx$. In time-step $t$, the quantization function for the $\bx$ can be expressed as:
\begin{equation}
    \hat{\bx} =\Phi(\left\lfloor{\frac{\bx}{s_{t}}}\right\rceil+z_{t}, 0,2^{b}-1).
    \label{fsc}
\end{equation}
To be noted, the calibration target is also aligned with the output of the Temporal Information Block. For a more specific estimation method of activation ranges, since the solution space within the finite set is limited, we find the Min-max~\cite{nagel2021white} method can achieve satisfactory results with high efficiency (more evidence in Sec.~\ref{tfc}). 
\section{Experiments}  
\subsection{Implementation details} 
\noindent\textbf{Models and datasets.} In this section, we conduct image generation experiments to evaluate the proposed TFMQ-DM framework on various diffusion models: pixel-space diffusion model DDPM~\cite{ho2020denoising} for unconditional image generation, latent-space diffusion model LDM~\cite{rombach2022ldm} for unconditional image generation and class-conditional image generation. We also apply our work to Stable Diffusion-v1-4 for text-guided image generation. In our experiments, We use seven standard benchmarks: CIFAR-10 $32 \times 32$~\cite{krizhevsky2009learning}, LSUN-Bedrooms $256 \times 256$~\cite{yu2016lsun}, LSUN-Churches $256 \times 256$~\cite{yu2016lsun}, CelebA-HQ $256 \times 256$~\cite{karras2018progressive}, ImageNet $256 \times 256$~\cite{deng2009imagenet}, FFHQ $256 \times 256$~\cite{karras2019stylebased} and MS-COCO~\cite{lin2015microsoft}.

\vspace{1em}
\noindent\textbf{Quantization settings.} We use channel-wise quantization for weights and layer-wise quantization for
activations, as it is a common practice. In our experimental setup, we employ BRECQ~\cite{li2021brecq} and AdaRound~\cite{nagel2020adaround}. Drawing from empirical insights derived from conventional model quantization practices~\cite{10.1007/978-3-319-46493-0_32, NIPS2015_ae0eb3ee}, we maintain the input and output layers of the model in full precision. Calibration sets, integral to our methodology, are generated through full-precision diffusion models, mirroring the approach outlined in Q-Diffusion~\cite{li2023qdiffusion}. Moreover, in weight quantization, we reconstruct quantized weights for $20$k iterations with a mini-batch size of 32 for DDPM and LDM, and 8 for Stable Diffusion. In activation quantization, we utilize EMA~\cite{jacob2018quantization} to estimate the ranges of activations with a mini-batch size of 16 on all models. More details can be found in the appendix.

\vspace{1em}
\noindent\textbf{Evaluation metrics.} For each experiment, we evaluate the performance of diffusion models with Fréchet Inception Distance (FID)~\cite{heusel2018gans}. In the case of LDM and Stable Diffusion experiments, we also include sFID~\cite{salimans2016improved}, which better captures spatial relationships than FID. For ImageNet and CIFAR-10 experiments, we additionally provide Inception Score (IS)~\cite{salimans2016improved} as a reference metric. Further, in the context of Stable Diffusion experiments, we extend our evaluation to include the compatibility of image-caption pairs, employing the CLIP score~\cite{hessel2022clipscore}. The ViT-B/32 is used as the backbone when computing the CLIP score. To ensure consistency in the reported outcomes,  all results are derived from our implementation or from other papers, where experiments are conducted under conditions consistent with ours. More specifically, in the evaluation process of each experiment, we sample 50k images from DDPM or LDM, or 30k images from Stable-Diffusion. All experiments are conducted utilizing one H800 GPU and implemented with
the PyTorch framework~\cite{paszke2019pytorch}.

\subsection{Main Results}
 
\textbf{Unconditional image generation.} 
\begin{table*}[!ht]\setlength{\tabcolsep}{5pt}
  \centering
  \caption{Quantization results for unconditional image generation with LDM-4 on LSUN-Bedrooms $256\time 256$, FFHQ $256\time 256$ and CelebA-HQ $256 \times 256$, LDM-8 on LSUN-Churches $256 \times 256$. * represents our implementation according to open-source codes and \dag means directly rerunning open-source codes.} 
  \resizebox{\linewidth}{!}{
\begin{tabular}{lcllllllll}
    \toprule
    \multicolumn{1}{c}{\multirow{2}{*}{\textbf{Methods}}} & \multicolumn{1}{c}{\multirow{2}{*}{\textbf{Bits (W/A)}}} & \multicolumn{2}{c}{\textbf{LSUN-Bedrooms $256 \times 256$}} & \multicolumn{2}{c}{\textbf{LSUN-Churches $256 \times 256$}} & \multicolumn{2}{c}{\textbf{CelebA-HQ $256 \times 256$}} & \multicolumn{2}{c}{\textbf{FFHQ $256 \times 256$}}\\ \cmidrule(r){3-4} \cmidrule(r){5-6} \cmidrule(r){7-8} \cmidrule(r){9-10}
    & & FID$\downarrow$ & sFID$\downarrow$ & FID$\downarrow$ & sFID$\downarrow$ & FID$\downarrow$ & sFID$\downarrow$ & FID$\downarrow$ & sFID$\downarrow$\\
    \midrule
    Full Prec. & 32/32 & 2.98 &  7.09 & 4.12 & 10.89 & 8.74 & 10.16 & 9.36 & 8.67\\
    \midrule
     PTQ4DM*~\cite{shang2022ptq4dm} & 4/32 & 4.83 &  7.94 & 4.92 & 13.94 & 13.67 & 14.72 & 11.74 & 12.18\\
    Q-Diffusion\dag~\cite{li2023qdiffusion} & 4/32 & 4.20 &  7.66 & 4.55 & 11.90 & 11.09 & 12.00 & 11.60 & 10.30\\
    PTQD*~\cite{he2023ptqd} & 4/32 & 4.42 &  7.88 & 4.67 & 13.68 & 11.06 & 12.21 & 12.01 & 11.12\\
    \rowcolor[gray]{0.9}TFMQ-DM~(Ours) & 4/32 & \textbf{3.60~\bl{(-0.60)}} &  \textbf{7.61~\bl{(-0.05)}} & \textbf{4.07~\bl{(-0.48)}} & \textbf{11.41~\bl{(-0.49)}} & \textbf{8.74~\bl{(-2.32)}} & \textbf{10.18~\bl{(-1.82)}} & \textbf{9.89~\bl{(-1.71)}} & \textbf{9.06~\bl{(-1.24)}}\\
    \midrule
    
    PTQ4DM*~\cite{shang2022ptq4dm} & 8/8 & 4.75 &  9.59 & 4.80 & 13.48 & 14.42 & 15.06 & 10.73 & 11.65\\
    Q-Diffusion\dag~\cite{li2023qdiffusion} & 8/8 & 4.51 &  8.17 & 4.41 & 12.23 & 12.85 & 14.16 & 10.87 & 10.01\\
    PTQD~\cite{he2023ptqd} & 8/8 & 3.75 &  9.89 & 4.89* & 14.89* & 12.76* & 13.54* & 10.69* & 10.97*\\
    \rowcolor[gray]{0.9}TFMQ-DM~(Ours) & 8/8 & \textbf{3.14~\bl{(-0.61)}} &  \textbf{7.26~\bl{(-0.91)}} & \textbf{4.01~\bl{(-0.40)}} & \textbf{10.98~\bl{(-1.25)}} & \textbf{8.71~\bl{(-4.05)}} & \textbf{10.20~\bl{(-3.34)}} & \textbf{9.46~\bl{(-1.23)}} & \textbf{8.73~\bl{(-1.28)}}\\
    \midrule
    
    PTQ4DM~\cite{shang2022ptq4dm} & 4/8 & 20.72 &  54.30 & 4.97* & 14.87* & 17.08* & 17.48* & 11.83* & 12.91*\\
    Q-Diffusion\dag~\cite{li2023qdiffusion} & 4/8 & 6.40 &  17.93 & 4.66 & 13.94 & 15.55 & 16.86 & 11.45 & 11.15\\
    PTQD~\cite{he2023ptqd} & 4/8 & 5.94 &  15.16 & 5.10* & 13.23* & 15.47* & 17.38* & 11.42* & 11.43*\\
    \rowcolor[gray]{0.9}TFMQ-DM~(Ours) & 4/8 & \textbf{3.68~\bl{(-2.26)}} &  \textbf{7.65~\bl{(-7.51)}} & \textbf{4.14~\bl{(-0.52)}} & \textbf{11.46~\bl{(-1.77)}} & \textbf{8.76~\bl{(-6.71)}} & \textbf{10.26~\bl{(-6.60)}} & \textbf{9.97~\bl{(-1.45)}} & \textbf{9.14~\bl{(-2.01)}}\\
    \bottomrule
\end{tabular}}
  \vspaceundertab
    \label{tab:sota_ldm}
\vspaceundercaption
\end{table*}
In the experiments conducted on the LDM, we maintain the same experimental settings as presented in ~\cite{rombach2022ldm}, including the number of steps, variance schedule, and classifier-free guidance scale  (denoted by eta and cfg in the following, respectively). As shown in Tab.~\ref{tab:sota_ldm}, the FID performance differences relative to the full precision (FP) model are all within 0.7 for all settings. Specifically, on the CelebA-HQ $256 \times 256$ dataset, our method exhibits a FID reduction of 6.71 and a sFID reduction of 6.60 in the w4a8 setting compared to the current state-of-the-art (SOTA). It is noticeable that existing methods, whether in 4-bit or 8-bit, show significant performance degradation when compared to the FP model on face datasets like CelebA-HQ $256 \times 256$ and FFHQ $256 \times 256$, whereas our TFMQ-DM shows almost no performance degradation compared to the FP model. Importantly, our method achieves significant performance improvement on the LSUN-Bedrooms $256 \times 256$ in the w4a8 setting, with FID and sFID reductions of 2.26 and 7.51 compared to PTQD~\cite{he2023ptqd}, respectively. Regarding LDM-8 on LSUN-Churches $256 \times 256$, we attribute the moderate improvement, compared to other datasets. We believe that the use of the LDM-8 model with a downsampling factor of 8 may be more quantization-friendly. Existing methods have already achieved satisfactory results on this dataset. Nonetheless, our method still approaches the performance of the FP model more closely compared to existing methods. Besides the experiments on LDM, We have also conducted experiments with DDPM on CIFAR-10 $32\times 32$, which can be found in the appendix.

\vspace{1em}
\noindent\textbf{Class-conditional image generation.}
\begin{table}[t]\setlength{\tabcolsep}{1pt}
  \centering
  \caption{Quantization results for unconditional image generation with class-conditional image generation with LDM-8 on ImageNet $256 \times 256$.} 
  \resizebox{\linewidth}{!}{
\begin{tabular}{lclll}
    \toprule
    \multicolumn{1}{c}{\multirow{2}{*}{\textbf{Methods}}} & \multicolumn{1}{c}{\multirow{2}{*}{\textbf{Bits (W/A)}}} & \multicolumn{3}{c}{\textbf{ImageNet $256 \times 256$}} \\ \cmidrule(r){3-5}
    & & IS$\uparrow$ & FID$\downarrow$ & sFID$\downarrow$\\
    \midrule
    Full Prec. & 32/32 & 235.64 & 10.91 & 7.67 \\
    \midrule
     PTQ4DM~\cite{shang2022ptq4dm} & 4/32 & - & - & - \\
    Q-Diffusion*~\cite{li2023qdiffusion} & 4/32 & 213.56 & 11.87 & 8.76 \\
    PTQD\dag~\cite{he2023ptqd} & 4/32 & 201.78 & 11.65 & 9.06 \\
    \rowcolor[gray]{0.9}TFMQ-DM~(Ours) & 4/32 & \textbf{223.81~\gre{(+10.25)}} & \textbf{10.50~\bl{(-1.15)}} & \textbf{7.98~\bl{(-0.78)}}\\
    \midrule
    
    PTQ4DM~\cite{shang2022ptq4dm} & 8/8 & 161.75 & 12.59 & - \\
    Q-Diffusion*~\cite{li2023qdiffusion} & 8/8 & 187.65 & 12.80 & 9.87 \\
    PTQD~\cite{he2023ptqd} & 8/8 & 153.92 & 11.94 & 8.03\\
    \rowcolor[gray]{0.9}TFMQ-DM~(Ours) & 8/8 & \textbf{198.86~\gre{(+11.21)}} & \textbf{10.79~\bl{(-1.15)}} & \textbf{7.65~\bl{(-0.38)}}\\
    \midrule
    
    PTQ4DM~\cite{shang2022ptq4dm} & 4/8 & - & - & - \\
    Q-Diffusion*~\cite{li2023qdiffusion} & 4/8 & 212.51 & 10.68 & 14.85 \\
    PTQD~\cite{he2023ptqd} & 4/8 & 214.73 & 10.40 & 12.63 \\
    \rowcolor[gray]{0.9}TFMQ-DM~(Ours) & 4/8 & \textbf{221.82~\gre{(+7.09)}} & \textbf{10.29~\bl{(-0.11)}} & \textbf{7.35~\bl{(-5.28)}}\\
    \bottomrule
\end{tabular}}
    \label{tab:sota_class}
\vspaceundercaption
\end{table}
On the ImageNet $256 \times 256$ dataset, we employed a denoising process with 20 iterations, setting eta and cfg to 0.0 and 3.0, respectively. Compared to PTQD, our method achieved a FID reduction of 1.15 on both w4a32 and w8a8. Simultaneously, in the w4a8 setting, sFID decreased by 5.28. Under the same conditions, we observed an improvement of over 7 in IS. Particularly noteworthy is that, across various quantization settings, our method consistently achieved lower FID compared to the FP model.

\vspace{1em}
\noindent\textbf{Text-guided image generation.}\begin{table}[t!]\setlength{\tabcolsep}{1pt}
  \centering
  \caption{Quantization results for text-guided image generation with Stable-Diffusion on MS-COCO captions.} 
  \resizebox{\linewidth}{!}{
\begin{tabular}[t!]{lcllll}
    \toprule
    \multicolumn{1}{c}{\multirow{2}{*}{\textbf{Methods}}} & \multicolumn{1}{c}{\multirow{2}{*}{\textbf{Bits (W/A)}}} & \multicolumn{3}{c}{\textbf{MS-COCO}} \\ \cmidrule(r){3-5}
    & & FID$\downarrow$ & sFID$\downarrow$ & CLIP$\uparrow$\\
    \midrule
    Full Prec. & 32/32 & 13.15 & 19.31 & 0.3146\\
    \midrule
    Q-Diffusion\dag~\cite{li2023qdiffusion} & 4/32 & 13.58 & 19.50 & 0.3143 \\
    \rowcolor[gray]{0.9}TFMQ-DM~(Ours) & 4/32 & \textbf{13.21~\bl{(-0.37)}} & \textbf{19.03~\bl{(-0.47)}} &
    \textbf{0.3144~\gre{(+0.0001)}}\\
    \midrule
    
     Q-Diffusion\dag~\cite{li2023qdiffusion} & 8/8 & 13.31 & 20.54 & 0.3134 \\
    \rowcolor[gray]{0.9}TFMQ-DM~(Ours) & 8/8 & \textbf{13.09~\bl{(-0.22)}} & \textbf{19.91~\bl{(-0.63)}} &
    \textbf{0.3134~\gre{(+0.0000)}}\\
    \midrule
    
     Q-Diffusion\dag~\cite{li2023qdiffusion} & 4/8 & 14.49 & 20.43 & 0.3121 \\
    \rowcolor[gray]{0.9}TFMQ-DM~(Ours) & 4/8 & \textbf{13.36~\bl{(-1.13)}} & \textbf{20.14~\bl{(-0.29)}} &
    \textbf{0.3128~
\gre{(+0.0007)}}\\
    \bottomrule
\end{tabular}
}
  \vspaceundertab
    \label{tab:sota_text}
\vspaceundercaption
\end{table}
In this experiment, we sample high-resolution images of $512 \times 512$ pixels with 50 denoising steps and fix cfg to the default 7.5 in Stable Diffusion as the trade-off between sample quality and diversity. In Tab.~\ref{tab:sota_text}, compared to the current SOTA Q-Diffusion, our approach achieves an FID reduction of 1.13 on w4a8. Simultaneously, our FID on w8a8 and sFID on w4a32 are even lower than those of the full precision model. However, existing metrics fail to adequately assess the semantic consistency of generated images. Nevertheless, based on the images generated in the appendix, our method produces higher-quality images with more realistic details, better demonstrating semantic information. Furthermore, our generated images closely approximate the effects of FP model.

\subsection{Ablation Study}\label{tfc}
To evaluate the effectiveness of each proposed method, we perform a thorough ablation study on the LSUN-Bedrooms $256 \times 256$ dataset with w4a8 quantization, utilizing the LDM-4 model with a DDIM sampler, as outlined in Tab.~\ref{tab:ablation}. We begin the assessment with a baseline BRECQ~\cite{li2021brecq}, which shows ineffective in denoising images when operating with 4-bit quantization. Additional ablation experiments can be found in the appendix.
\begin{table}[t]\setlength{\tabcolsep}{5pt}
  \centering
  \caption{The effect of different methods proposed in the paper on LSUN-Bedrooms $256 \times 256$.} 
  \resizebox{\linewidth}{!}{
  \begin{tabular}{lcll}
    \toprule
    \textbf{Methods} & \textbf{Bits (W/A)} & FID$\downarrow$ & sFID$\downarrow$ \\
    \midrule
    Full Prec. & 32/32 & 2.98 & 7.09 \\
    \midrule
    BRECQ~\cite{li2021brecq} (Baseline) & 4/8  & 9.36 & 22.73 \\
    +TIAR & 4/8  & 4.84 & 9.29\\
    +FSC & 4/8  & 6.07 & 11.31 \\
    \rowcolor[gray]{0.9}+TFMQ-DM (TIAR + FSC) & 4/8 & \textbf{3.68~\bl{(-5.68)}} & \textbf{7.65~\bl{(-15.08)}} \\
    \bottomrule
\end{tabular}
}
  \vspaceundertab
    \label{tab:ablation}
\vspaceundercaption
\end{table}

\begin{figure}[!ht]
    \centering
     \setlength{\abovecaptionskip}{0.2cm}
    \includegraphics[width=0.35\textwidth]{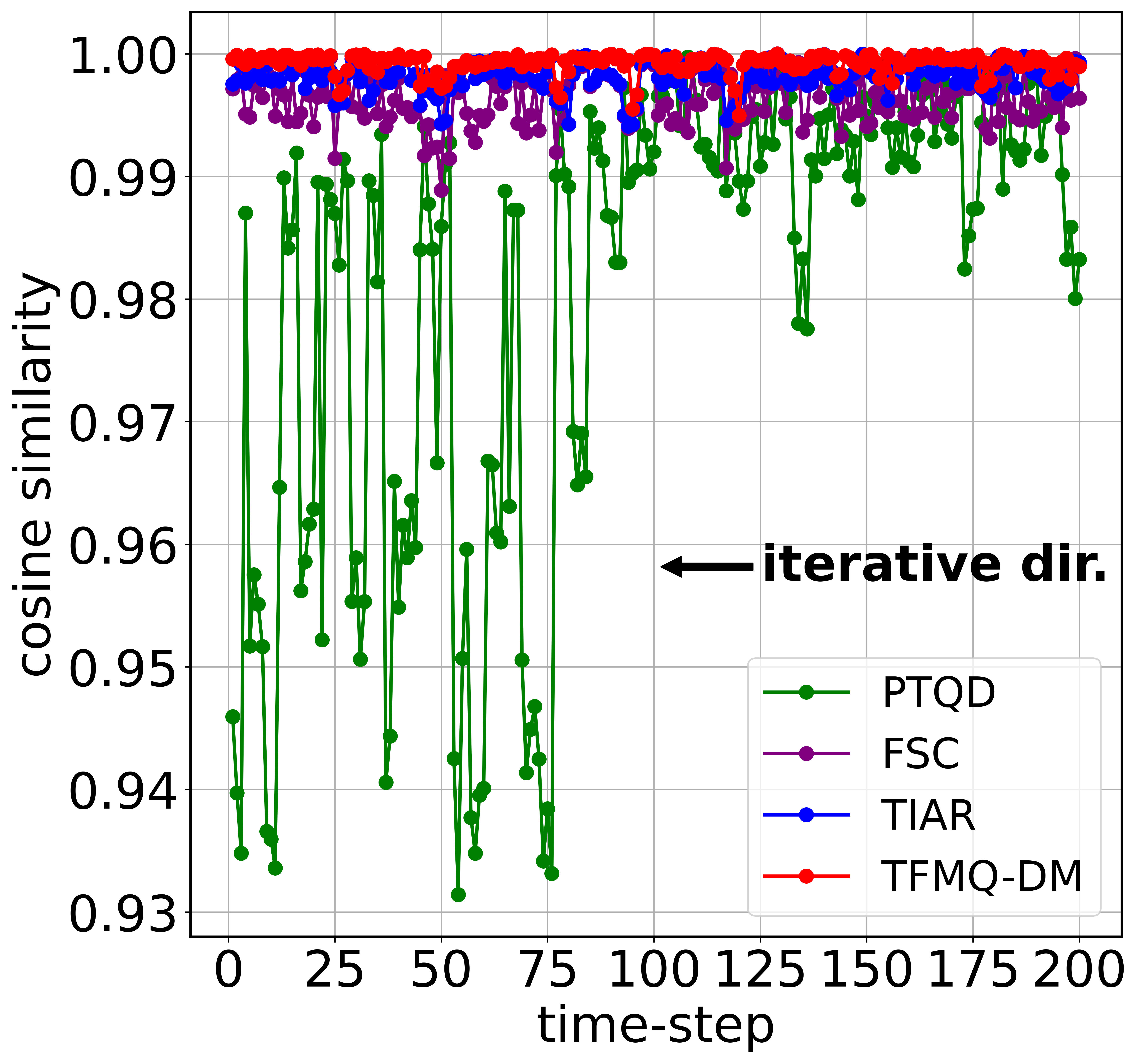}
     \caption{Temporal feature errors across different PTQ methods.}
    \label{ablation}
    \vspaceundercaption
\end{figure}

\vspace{1em}
\noindent\textbf{Effect of TIAR.} It can be observed that, compared to the baseline, our TIAR method reduces FID and sFID by 4.52 and 13.44, respectively. Furthermore, As shown in Fig.~\ref{ablation}, our method's temporal feature disturbance is significantly far weaker than existing SOTA PTQD. This indicates the effectiveness of our method in maintaining temporal information contained in temporal features. Further details of the effects are presented in the appendix.

\vspace{1em}
\noindent\textbf{Effect of FSC.} Our FSC method has also achieved remarkably positive results. Compared to the baseline, it reduces FID and sFID by 3.29 and 11.42, respectively. Compared to PTQD, from Fig.~\ref{ablation}, our method's temporal feature error is significantly smaller than that of PTQD. Since we employ layer-wise quantization for activations, this introduces less than one percent of additional parameters. The inference time overhead incurred by switching different step sizes and zero points during multiple time-steps inference is negligible, as detailed in the appendix.

\vspace{1em}
\noindent\textbf{Efficiency of FSC.} For FSC, there are various methods to evaluate the range of activations to determine the optimal step size. We try several methods and assess the GPU time consumed during calibration, as detailed in Tab.~\ref{tab:cali_method}. Since the improvements achieved by these methods in model performance are similar, we opt for the simplest and most efficient Min-max~\cite{nagel2021white} method as our specific calibration strategy, striking a balance between calibration time and effectiveness. Notably, we have found that PTQD and Q-Diffusion cost 4.68 and 5.29 GPU hours in their PTQ methods under w4a8 quantization on LSUN-Bedrooms $256\times 256$, respectively. However, our framework only spends 2.32 GPU hours (as shown in Tab.~\ref{tab:cali_method}).

Furthermore, for the only learning-based method in the table, LSQ~\cite{esser2020lsq}, which is one of the most commonly used methods in previous works~\cite{he2023ptqd, shang2022ptq4dm, li2023qdiffusion}, we observe that it did not outperform other methods and, in some cases, performs even worse. We speculate that the main reason might be that, for a fixed time-step, the calibration data used to determine the quantization parameters is relatively limited compared to all calibration data. In such cases, LSQ may struggle to learn an optimal set of parameters.
\begin{table}[t]\setlength{\tabcolsep}{1pt}
  \centering
  \caption{Different calibration methods for FTFC. We note $T$ as calibration GPU hours in the table.} 
  \resizebox{\linewidth}{!}{
\begin{tabular}{lclll}
    \toprule
    \textbf{TSC Methods} & \textbf{Bits (W/A)} & FID$\downarrow$ & sFID$\downarrow$ & T (hours)\\
    \midrule
    FIAR & 8/32 & 2.84 & 7.08 & 2.18\\
    \cdashline{1-5}
    +LSQ~\cite{esser2020lsq} & 8/8 & 3.17 & 7.18 & 2.48 \\
    +KL-divergence & 8/8 & 3.27 & 7.32 & 19.67 \\
    +Percentile & 8/8 & 3.34 & 7.41 & 12.00 \\
    +MSE & 8/8 & \textbf{3.12} & \textbf{7.12} & 8.89 \\
    \rowcolor[gray]{0.9}+Min-max~\cite{nagel2021white} & 8/8 & 3.14~\gre{(+0.02)} & 7.26~\gre{(+0.14)} & \textbf{0.12~\bl{(-2.36)}} \\
    \midrule
    FIAR & 4/32 & 3.04 & 7.16 & 2.20\\
    \cdashline{1-5}
    +LSQ~\cite{esser2020lsq} & 4/8 & 3.69 & 7.48 & 2.57 \\
    +KL-divergence & 4/8 & 3.94 & \textbf{7.42} & 19.65 \\
    +Percentile & 4/8 & 3.74 & 8.02 & 12.04 \\
    +MSE & 4/8 & \textbf{3.62} & 7.48 & 8.89 \\
    \rowcolor[gray]{0.9}+Min-max~\cite{nagel2021white} & 4/8 & 3.68~\gre{(+0.06)} & 7.65~\gre{(+0.23)} & \textbf{0.12~\bl{(-2.45)}} \\
    \bottomrule
\end{tabular}}
  \vspaceundertab
    \label{tab:cali_method}
\vspaceundercaption
\end{table}
\section{Conclusion \& Discussions of Limitations} 
This research explores the application of quantization for accelerating diffusion models. In this work, we identify a novel and significant problem, namely temporal feature disturbance, in the quantization of diffusion models. We conducted a detailed analysis of the root causes of this problem and introduced our TFMQ-DM quantization framework. In 4-bit quantization on extensive datasets and different diffusion models, this framework exhibits minimal performance degradation compared to the FP model and speedup quantization time.

However, besides temporal features, we have also found the textual features introduced in Stable Diffusion have a physical meaning and influence the generation effect. Nevertheless, these textual features are not taken into consideration in current methods. Secondly, temporal feature maintenance is suitable for both PTQ and QAT scenarios. But in this study, our main focus is on the PTQ setting and achieving 4-bit quantization. The extension to the QAT setting should be conducted in the future to achieve lower-bit quantization and further performance enhancements.
{\small
\bibliographystyle{ieeenat_fullname}
\bibliography{main}
}
\newpage
\appendix
\begin{center}{\bf \Large Appendix}\end{center}\vspace{-2mm}
\renewcommand{\thetable}{\Roman{table}}
\renewcommand{\thefigure}{\Roman{figure}}
\setcounter{table}{0}
\setcounter{figure}{0}

\Crefname{appendix}{Appendix}{Appendixes}
Note: All references to figures or tables identified by Arabic numerals point to the corresponding figures or tables in the main text.
\section{More Implementation Details}
In our reconstruction and calibration, apart from the Temporal Information Block proposed by us, the partitioning of the remaining network components remains consistent with PTQD~\cite{he2023ptqd} and Q-Diffusion~\cite{li2023qdiffusion} (\emph{i.e.}, Residual Bottleneck Blocks, Attention Blocks, and the remaining layers). Specifically, for the reconstruction of the Residual Bottleneck Block, we freeze the quantization parameters of the \verb|embedding layer|, and these parameters are only tuned in the reconstruction within the Temporal Information Block.

Additionally, the quantization settings are kept consistent with Q-Diffusion and PTQD.



\section{Activation Range Variations in Finite Set}
\begin{figure*}[t]
    \centering
    \setlength{\abovecaptionskip}{0.2cm}
    \includegraphics[width=\textwidth]{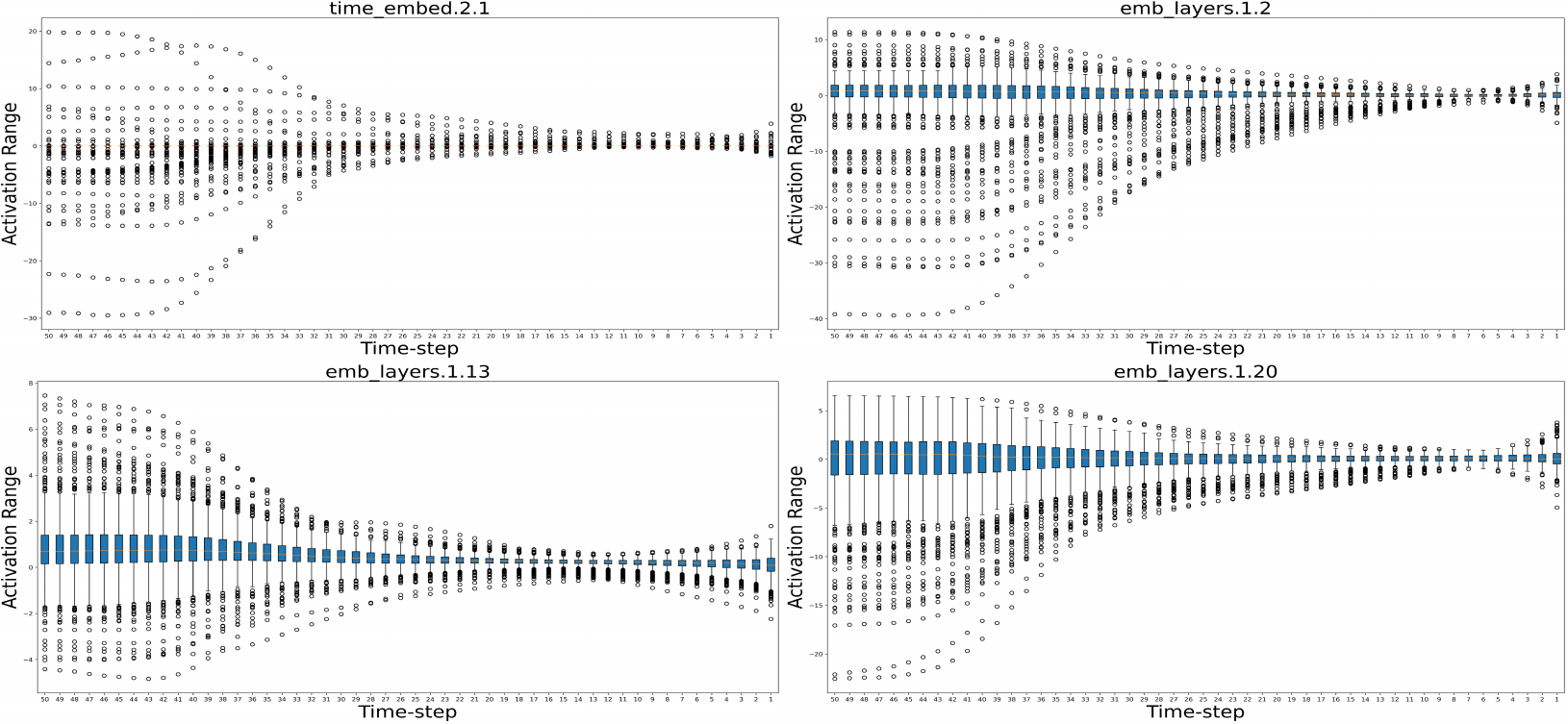}
     \caption{Activation ranges within sampling data-unrelated components for LDM-4 on LSUN-Bedrooms $256\times 256$  with 50 denoising steps. We randomly select 4 linear or convolutional layers' activations in these components to demonstrate the range variation.}
    \label{act_range}
    \vspaceundercaption
\end{figure*}

We analyze activation value ranges across all time steps in sampling data-unrelated components, \emph{e.g.}, \verb|time embed| and \verb|embedding layers| for diffusion models. In Fig.~\ref{act_range}, it is evident that activation ranges vary notably among different time steps within these components. This observation suggests that the activation ranges within the same layer undergo considerable changes with varying time steps. Fortunately, the activations in the Time Information Block belong to a finite set, providing us the opportunity to conduct an accurate calibration for each time step.

\section{Inappropriate Calibration Target} 
In this part, we further conduct experiments to provide the clues that the inappropriate reconstruction target also results in an inappropriate calibration. In the previous works, they calibrate the \verb|embedding layers| along with the corresponding Residual Bottleneck Blocks. On the contrary, we freeze the quantized parameters of the \verb|embedding layers| during the calibration process with a simple Min-max~\cite{nagel2021white} initialization, which separates the calibration of \verb|embedding layers| as alone. The experimental results in Tab.~\ref{tab:cali_comp} demonstrate that without calibrating these layers inside the Residual Bottleneck Block can achieve better results. This confirms that the inappropriate calibration target leads to the suboptimal tuning of the quantization parameters.
\begin{table}[!ht]\setlength{\tabcolsep}{10pt}
  \centering
  \caption{FID and sFID on LSUN-Bedrooms $256\times256$~\cite{yu2016lsun} for LDM-4. Prev represents BRECQ, the same as Tab.~\re{1}. Freeze denotes our trial here.} 
  \resizebox{0.9\linewidth}{!}{
  \begin{tabular}{lcll}
    \toprule
    \textbf{Methods} & \textbf{Bits (W/A)} & FID$\downarrow$ & sFID$\downarrow$ \\
    \midrule
    Full Prec. & 32/32 & 2.98 & 7.09 \\
    \midrule
    Prev & 8/8 & 7.51 &  12.54 \\
    \rowcolor[gray]{0.9}Freeze & 8/8 & \textbf{6.87~\bl{(-0.64)}} & \textbf{10.12~\bl{(-2.42)}} \\
    \midrule
    Prev & 4/8  & 9.36 & 22.73 \\
    \rowcolor[gray]{0.9}Freeze & 4/8 & \textbf{8.06~\bl{(-1.30)}} & \textbf{18.47~\bl{(-4.26)}} \\
    \bottomrule
\end{tabular}}
    \label{tab:cali_comp}
\end{table}

\section{Unconditional Image Generation on CIFAR-10}
In this section, we conduct more experiments for unconditional image generation on CIFAR-10 $32 \times 32$. As shown in Fig.~\ref{tab:sota_ddim}, our methods still achieve comprehensive improvements in terms of IS and FID compared to the existing SOTA. However, due to the lower resolution and simplicity of the images in this dataset, existing methods show minimal performance degradation, so the results we obtain may not be as pronounced.
\begin{table}[!ht]\setlength{\tabcolsep}{10pt}
  \centering
  \caption{Quantization results for unconditional image generation with DDIM on CIFAR-10 $32 \times 32$.} 
  \resizebox{0.9\linewidth}{!}{
\begin{tabular}{lcll}
    \toprule
    \multicolumn{1}{c}{\multirow{2}{*}{\textbf{Methods}}} & \multicolumn{1}{c}{\multirow{2}{*}{\textbf{Bits (W/A)}}} & \multicolumn{2}{c}{\textbf{CIFAR-10 $32 \times 32$}} \\ \cmidrule(r){3-4}
    & & IS$\uparrow$ & FID$\downarrow$ \\
    \midrule
    Full Prec. & 32/32 & 9.04 & 4.23 \\
    \midrule
     PTQ4DM*~\cite{shang2022ptq4dm} & 4/32 & 9.02 & 5.65 \\
    Q-Diffusion\dag~\cite{li2023qdiffusion} & 4/32 & 8.78 & 5.08 \\
    TDQ~\cite{so2023temporal} & 4/32 & - & - \\
    \rowcolor[gray]{0.9}TFMQ-DM~(Ours) & 4/32 & \textbf{9.14~\gre{(+0.12)}} & \textbf{4.73~\bl{(-0.35)}} \\
    \midrule
    
    PTQ4DM~\cite{shang2022ptq4dm} & 8/8 & 9.02 & 19.59 \\
    Q-Diffusion\dag~\cite{li2023qdiffusion} & 8/8 & 8.89 & 4.78 \\
    TDQ~\cite{so2023temporal} & 8/8 & 8.85 & 5.99\\
    \rowcolor[gray]{0.9}TFMQ-DM~(Ours) & 8/8 & \textbf{9.07~\gre{(+0.05)}} & \textbf{4.24~\bl{(-0.54)}} \\
    \midrule
    
    PTQ4DM*~\cite{shang2022ptq4dm} & 4/8 & 8.93 & 5.14 \\
    Q-Diffusion\dag~\cite{li2023qdiffusion} & 4/8 & 9.12 & 4.98 \\
    TDQ~\cite{so2023temporal} & 4/8 & - & - \\
    \rowcolor[gray]{0.9}TFMQ-DM~(Ours) & 4/8 & \textbf{9.13~\gre{(+0.01)}} & \textbf{4.78~\bl{(-0.20)}} \\
    \bottomrule
\end{tabular}}
  \vspaceundertab
    \label{tab:sota_ddim}
\end{table}

\section{Additional Effect of TIAR}
\begin{figure}[!ht]
    \centering
     \setlength{\abovecaptionskip}{0.2cm}
    \includegraphics[width=0.35\textwidth]{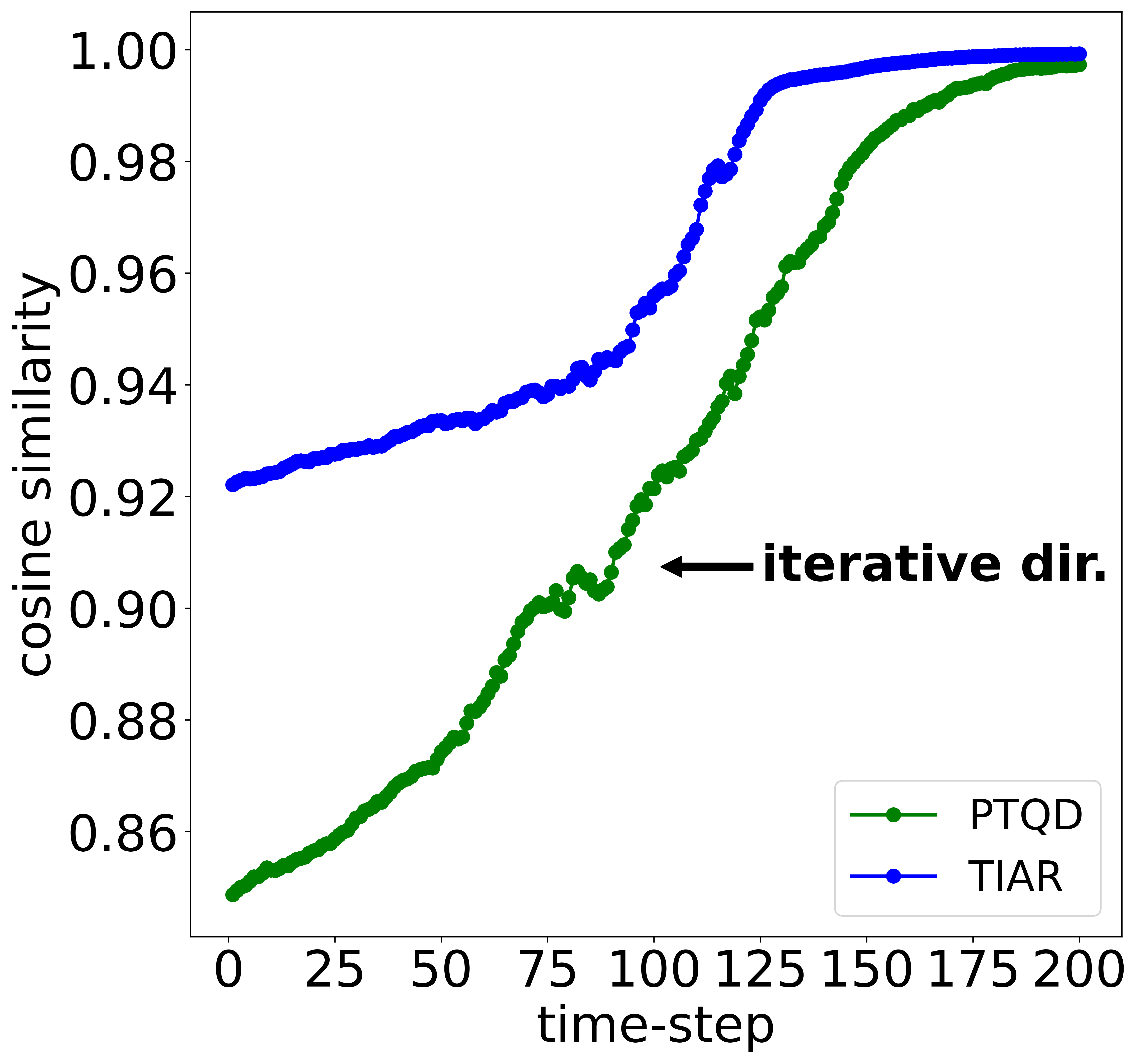}
     \caption{Cosine similarity of the Residual Bottleneck's outputs across different PTQ Methods.}
    \label{res_out}
    \vspaceundercaption
\end{figure}

As shown in Fig.~\re{5}, both of our proposed methods for LDM-4 on LSUN-Bedrooms $256\times 256$ significantly reduce temporal feature errors, thereby alleviating temporal feature disturbance to a great extent. In this section, we conduct a detailed analysis of the cosine similarity between the outputs of the $i^{\text{th}}$ Residual Bottleneck Blocks before and after quantization. We compare the results obtained with our TIAR and PTQD under w4a8 quantization, where $i=8$ and $T=200$ (the same as the settings in Fig.~\re{5}). From Fig.~\ref{res_out}, it can be observed that our approach significantly reduces output errors of the Residual Bottleneck Block compared to PTQD. However, it is essential to note that the error at this point involves the accumulated errors from multiple denoising iterations in diffusion models. Since Fig.~\re{5} is not subject to the impact of accumulated errors, the trends of the lines in the two graphs may exhibit slight differences.

\section{Inference Cost of TSC}
In this section, we assess the inference overhead of our TFMQ-DM method on real hardware, specifically the Intel$^\circledR$ Xeon$^\circledR$ Gold 6248R Processor. All floating-point and quantized operations are implemented using Intel's OpenVINO toolkit~\footnote{\href{https://www.intel.com/content/www/us/en/developer/tools/openvino-toolkit/overview.html}{OpenVINO toolkit}}. As illustrated in Table~\ref{tab:infer}, in comparison to the UNet quantized with the built-in w8a8 quantization method in the OpenVINO toolkit, our approach results in a memory overhead of less than $0.076\%$, yielding a $2.38\times$ acceleration compared to the original floating-point model. Moreover, our method introduces less than $0.5\%$ additional latency compared to the built-in w8a8 quantization in the OpenVINO toolkit.
\begin{table}[!ht]\setlength{\tabcolsep}{10pt}
  \centering
  \caption{Inference analysis of Stable Diffusion with 50 denoising time-steps on Intel CPU.} 
  \resizebox{\linewidth}{!}{
  \begin{tabular}{lcccc}
    \toprule
    \textbf{Methods} & \textbf{Bits (W/A)} & UNet Size (Mb) & Latency (s) & Speedup\\
    \midrule
    Full Prec. & 32/32 & 3278.81 & 81.01 & -\\
    \midrule
    OpenVINO & 8/8  & 821.15 & 33.93 & $2.39\times$\\
    \rowcolor[gray]{0.9}TFMQ-DM & 8/8 & 821.77 &  34.07 & $2.38\times$\\
    \bottomrule
\end{tabular}}
  \vspaceundertab
    \label{tab:infer}
\vspaceundercaption
\end{table}

\section{Study of Sampling with Advanced Samplers}
Apart from employing the DDIM sampler~\cite{songddim}, we also utilize a variant of DDPM~\cite{ho2020denoising} called PLMS~\cite{liu2022pseudo} on the CelebA-HQ $256\times 256$ dataset~\cite{karras2018progressive}. This better demonstrates the superiority of our TFMQ-DM framework compared to previous works. From Tab.~\ref{tab:sota_plms}, the introduced TFMQ-DM substantially reduces FID and sFID, surpassing PTQD by margins of 12.40 and 7.09, respectively.
\begin{table}[!ht]\setlength{\tabcolsep}{10pt}
  \centering
  \caption{Quantization results for unconditional image generation with PLMS on CelebA-HQ $256 \times 256$.} 
  \resizebox{0.9\linewidth}{!}{
\begin{tabular}{lcll}
    \toprule
    \multicolumn{1}{c}{\multirow{2}{*}{\textbf{Methods}}} & \multicolumn{1}{c}{\multirow{2}{*}{\textbf{Bits (W/A)}}} & \multicolumn{2}{c}{\textbf{CelebA-HQ $256 \times 256$}} \\ \cmidrule(r){3-4}
    & & FID$\downarrow$ & sFID$\downarrow$ \\
    \midrule
    Full Prec. & 32/32 & 8.92 & 10.42 \\
    \midrule
    Q-Diffusion~\cite{li2023qdiffusion} & 4/8 & 24.31 & 22.11 \\
    PTQD~\cite{he2023ptqd} & 4/8 & 21.08 & 17.38 \\
    \rowcolor[gray]{0.9}TFMQ-DM~(Ours) & 4/8 & \textbf{8.68~\bl{(-12.40)}} & \textbf{10.29~\bl{(-7.09)}} \\
    \bottomrule
\end{tabular}}
  \vspaceundertab
    \label{tab:sota_plms}
\end{table}

Additionally, we present experiments performed using the DPM++ solver~\cite{lu2023dpmsolver} on LSUN-Churches $256 \times 256$~\cite{yu2016lsun}. As illustrated in Tab.~\ref{tab:sota_dpm}, our framework consistently outperforms existing methods in terms of performance on this dataset with the DPM++ solver.
\begin{table}[!ht]\setlength{\tabcolsep}{10pt}
  \centering
  \caption{Quantization results for unconditional image generation with DPM++ on LSUN-Churches $256 \times 256$.} 
  \resizebox{0.9\linewidth}{!}{
\begin{tabular}{lcll}
    \toprule
    \multicolumn{1}{c}{\multirow{2}{*}{\textbf{Methods}}} & \multicolumn{1}{c}{\multirow{2}{*}{\textbf{Bits (W/A)}}} & \multicolumn{2}{c}{\textbf{LSUN-Churches $256 \times 256$}} \\ \cmidrule(r){3-4}
    & & FID$\downarrow$ & sFID$\downarrow$ \\
    \midrule
    Full Prec. & 32/32 & 4.12 & 10.55 \\
    \midrule
    Q-Diffusion~\cite{li2023qdiffusion} & 4/8 & 7.80 & 23.24 \\
    PTQD~\cite{he2023ptqd} & 4/8 & 7.45 & 22.74 \\
    \rowcolor[gray]{0.9}TFMQ-DM~(Ours) & 4/8 & \textbf{5.51~\bl{(-1.94)}} & \textbf{13.15~\bl{(-9.59)}} \\
    \bottomrule
\end{tabular}}
  \vspaceundertab
    \label{tab:sota_dpm}
\vspaceundercaption
\end{table}

\section{Comparison of Visualization Results}
\begin{figure*}[tp!]
   \centering
    \setlength{\abovecaptionskip}{0.2cm}
    \begin{subfigure}[tp!]{0.8\textwidth}
        \centering
        \includegraphics[width=\textwidth]{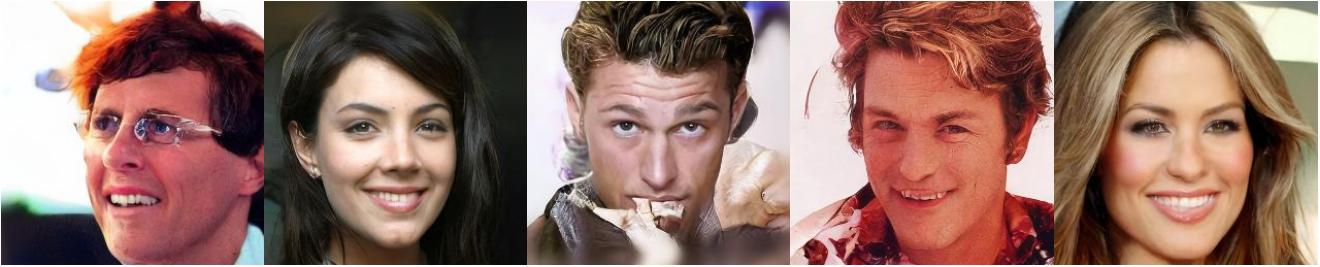}
        \subcaption{FP}
    \end{subfigure}
    \begin{subfigure}[tp!]{0.8\textwidth}
        \centering
        \includegraphics[width=\textwidth]{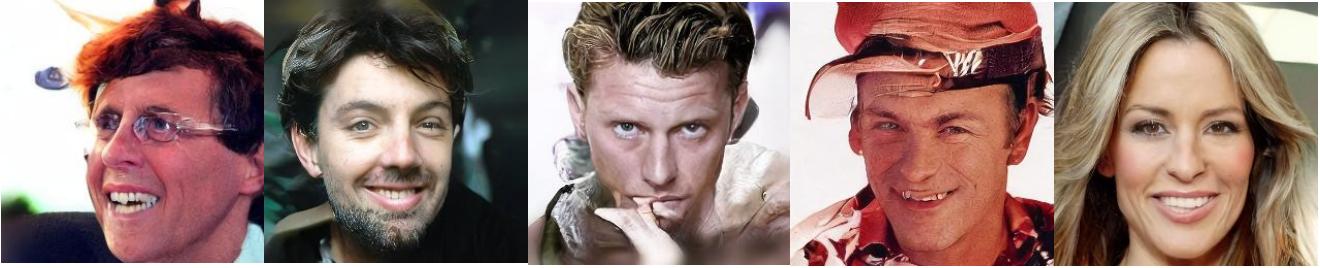}
        \subcaption{Q-Diffusion (w4a8)}
    \end{subfigure}
    \begin{subfigure}[tp!]{0.8\textwidth}
        \centering
        \includegraphics[width=\textwidth]{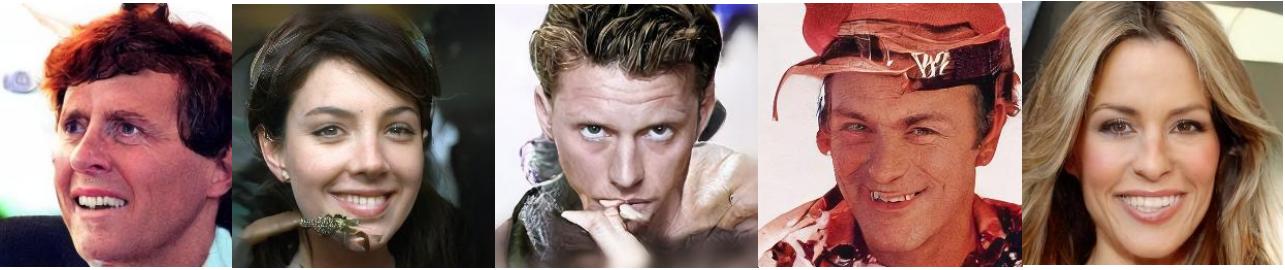}
        \subcaption{PTQD (w4a8)}
    \end{subfigure}
    \begin{subfigure}[tp!]{0.8\textwidth}
        \centering
        \includegraphics[width=\textwidth]{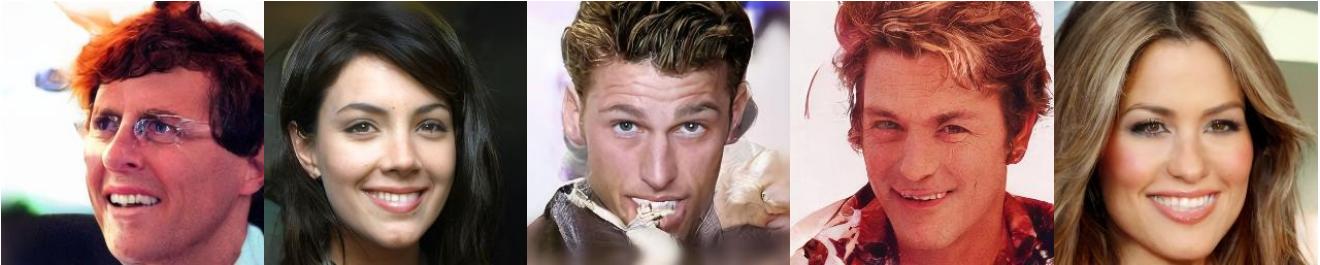}
        \subcaption{TFMQ-DM (w4a8)}
    \end{subfigure}
    \caption{Random samples from w4a8 quantized and full-precision LDM-4 on CelebA-HQ $256 \times 256$. The resolution of each sample is $256\times256$.}
    \label{first_celeba-hq}
\end{figure*}

\begin{figure*}[tp!]
   \centering
    \setlength{\abovecaptionskip}{0.2cm}
    \begin{subfigure}[tp!]{\textwidth}
        \centering
        \includegraphics[width=\textwidth]{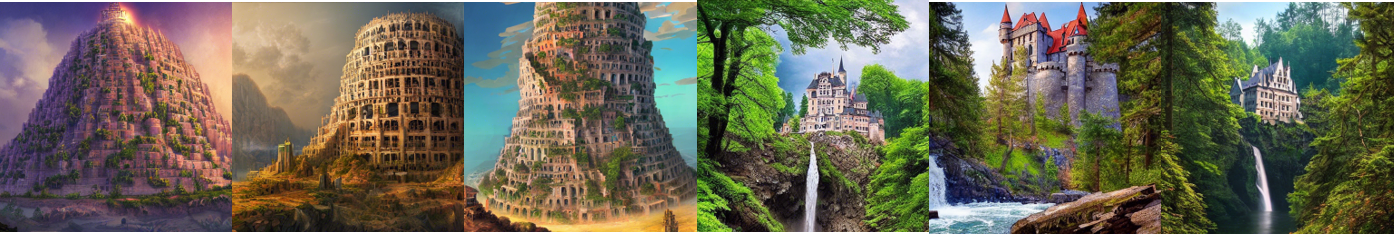}
        \subcaption{FP}
    \end{subfigure}
    \begin{subfigure}[tp!]{\textwidth}
        \centering
        \includegraphics[width=\textwidth]{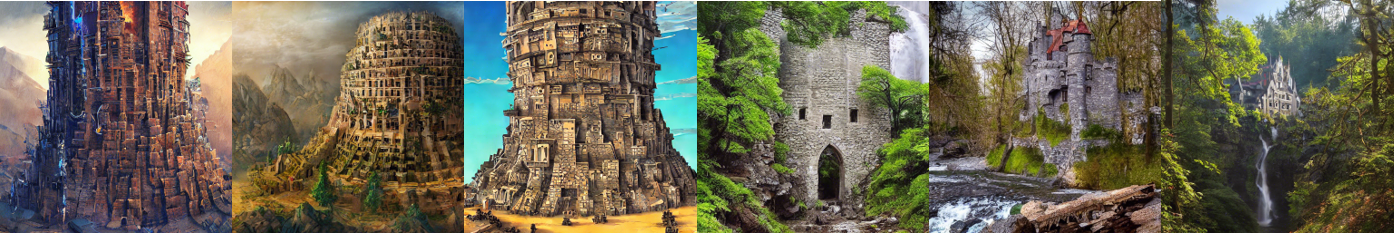}
        \subcaption{Q-Diffusion (w4a8)}
    \end{subfigure}
    \begin{subfigure}[tp!]{\textwidth}
        \centering
        \includegraphics[width=\textwidth]{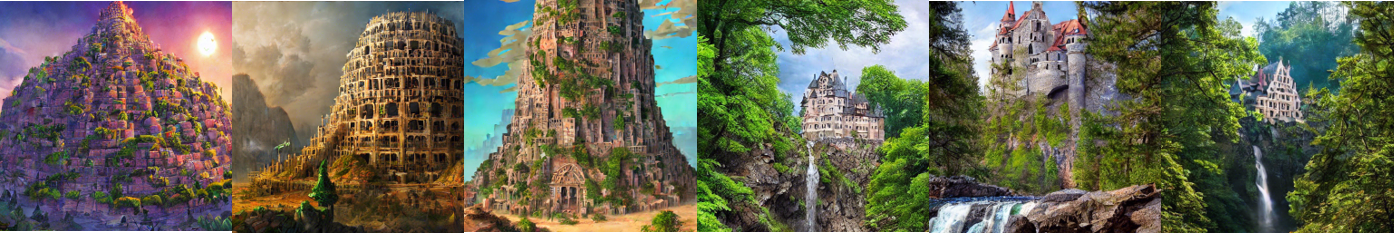}
        \subcaption{TFMQ-DM (w4a8)}
    \end{subfigure}
    \caption{Random samples from w4a8 quantized and full-precision Stable Diffusion. (Left) prompt: \textit{A digital illustration of the Babel tower, detailed, trending in artstation, fantasy vivid colors}. (Right) prompt: \textit{A beautiful castle beside a waterfall in the woods}. The resolution of each sample is $512\times 512$.}
    \label{last_sd}
\end{figure*}

Within this section, we present random samples derived from full-precision and w4a8 quantized diffusion models with a fixed random seed. These quantized models were created through our TFMQ-DM or previous state-of-the-art methods. The figures below illustrate the obtained results. As shown from Fig.~\ref{first_celeba-hq} to Fig.~\ref{last_sd}, our framework yields results that closely resemble those of the full-precision model, showcasing higher fidelity. Moreover, it excels in finer details, producing superior outcomes in some intricate aspects (zoom in to closely examine the relevant images).
\begin{figure*}[tp!]
   \centering
    \setlength{\abovecaptionskip}{0.2cm}
    \begin{subfigure}[tp!]{0.6\textwidth}
        \centering
        \includegraphics[width=\textwidth]{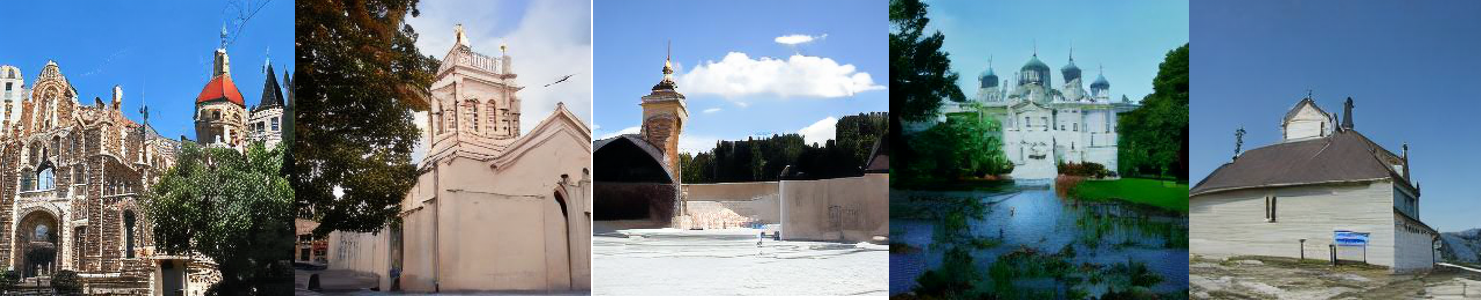}
        \subcaption{FP}
    \end{subfigure}
    \begin{subfigure}[tp!]{0.6\textwidth}
        \centering
        \includegraphics[width=\textwidth]{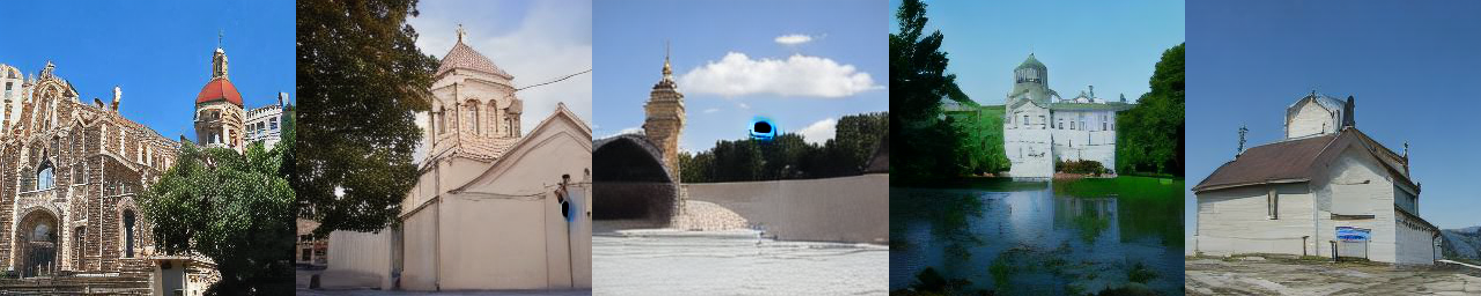}
        \subcaption{Q-Diffusion (w4a8)}
    \end{subfigure}
    \begin{subfigure}[tp!]{0.6\textwidth}
        \centering
        \includegraphics[width=\textwidth]{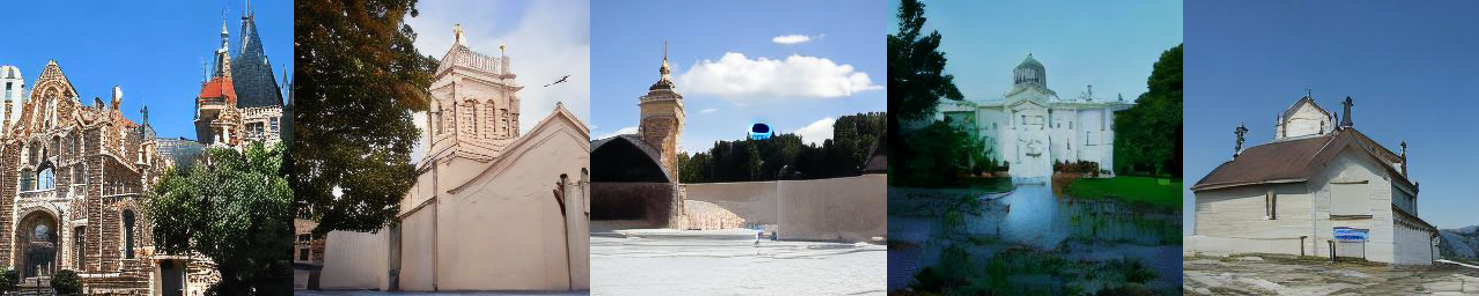}
        \subcaption{PTQD (w4a8)}
    \end{subfigure}
    \begin{subfigure}[tp!]{0.58\textwidth}
        \centering
        \includegraphics[width=\textwidth]{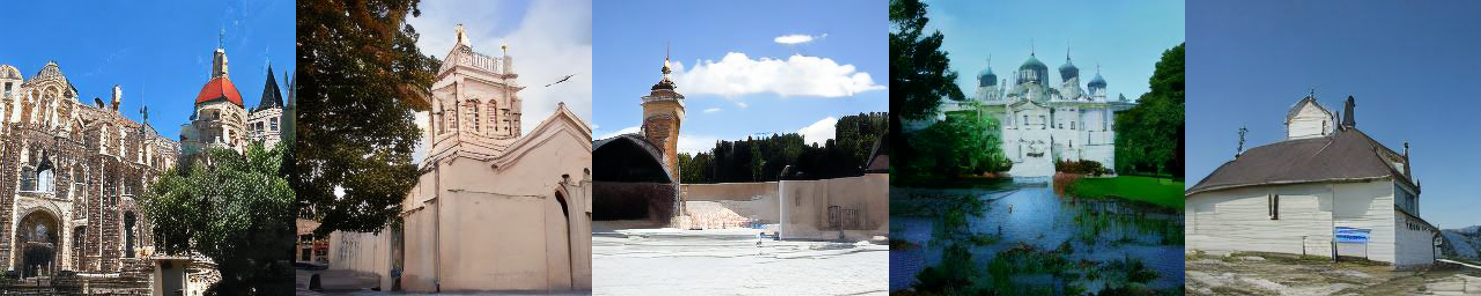}
        \subcaption{TFMQ-DM (w4a8)}
    \end{subfigure}
    \caption{Random samples from w4a8 quantized and full-precision LDM-8 on LSUN-Churches $256 \times 256$. The resolution of each sample is $256\times256$.}
\end{figure*}
\begin{figure*}[tp!]
   \centering
    \setlength{\abovecaptionskip}{0.2cm}
    \begin{subfigure}[tp!]{0.7\textwidth}
        \centering
        \includegraphics[width=\textwidth]{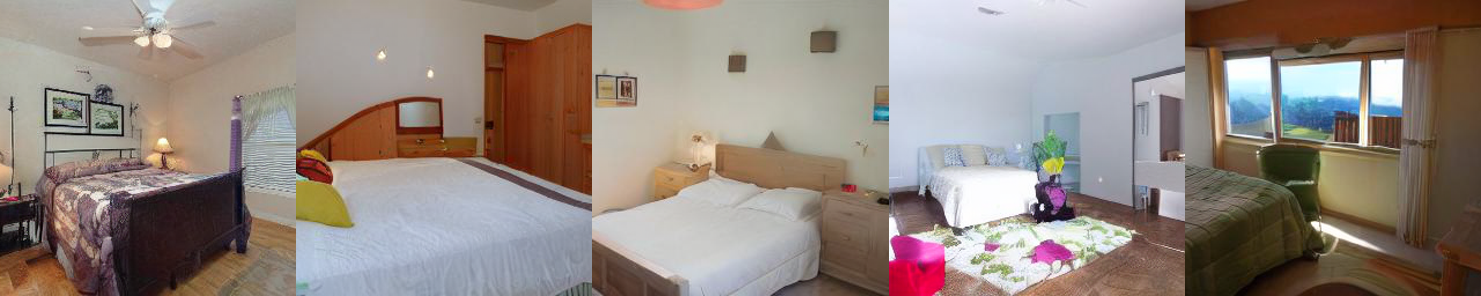}
        \subcaption{FP}
    \end{subfigure}
    \begin{subfigure}[tp!]{0.7\textwidth}
        \centering
        \includegraphics[width=\textwidth]{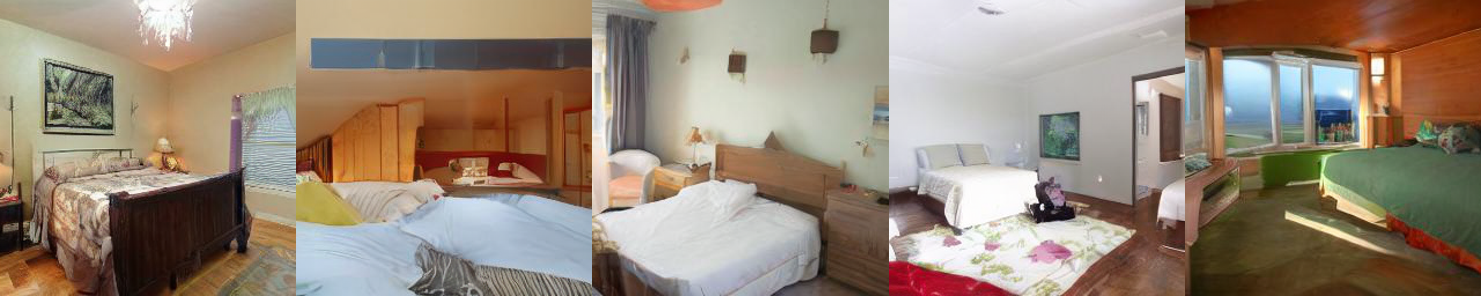}
        \subcaption{PTQD (w4a8)}
    \end{subfigure}
    \begin{subfigure}[tp!]{0.7\textwidth}
        \centering
        \includegraphics[width=\textwidth]{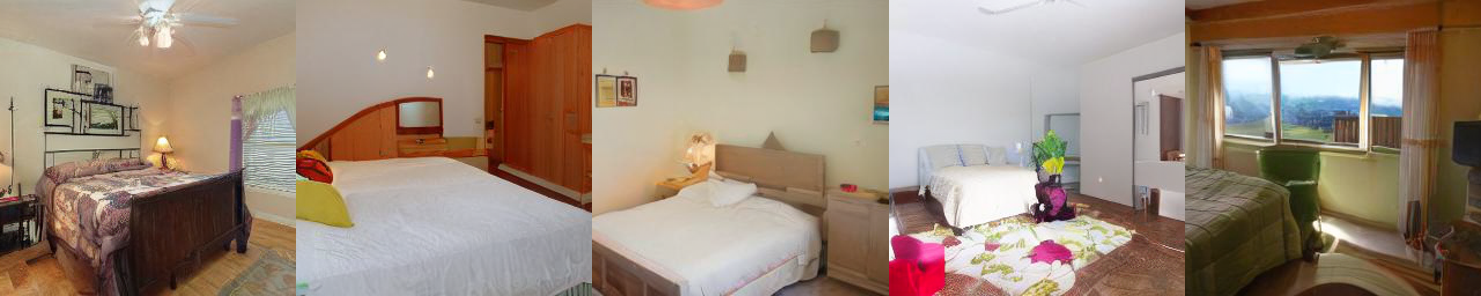}
        \subcaption{TFMQ-DM (w4a8)}
    \end{subfigure}
    \caption{Random samples from w4a8 quantized and full-precision LDM-4 on LSUN-Bedrooms $256 \times 256$. The resolution of each sample is $256\times256$.}
\end{figure*}
\begin{figure*}[tp!]
   \centering
    \setlength{\abovecaptionskip}{0.2cm}
    \begin{subfigure}[tp!]{0.55\textwidth}
        \centering
        \includegraphics[width=\textwidth]{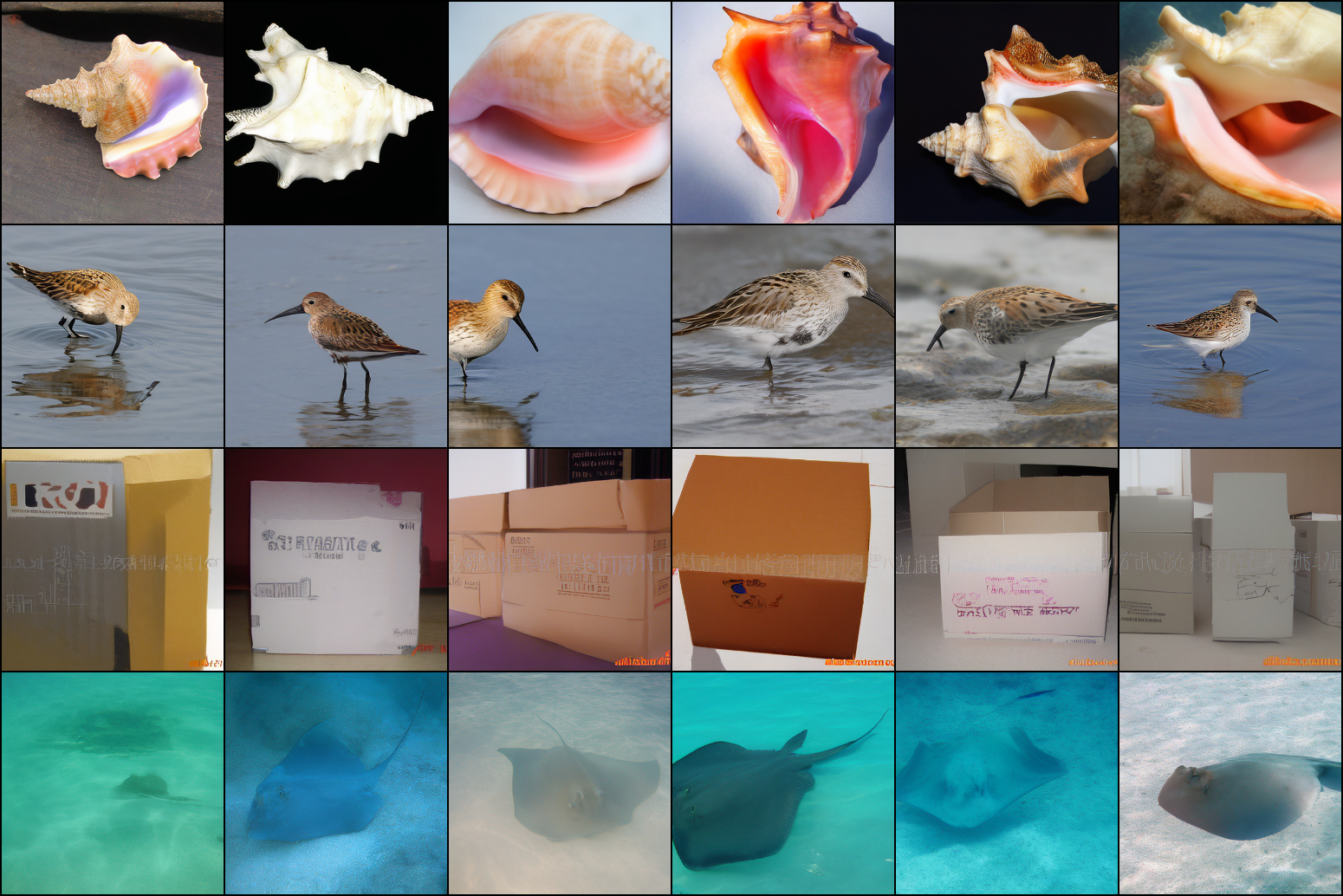}
        \subcaption{FP}
    \end{subfigure}
    \begin{subfigure}[tp!]{0.55\textwidth}
        \centering
        \includegraphics[width=\textwidth]{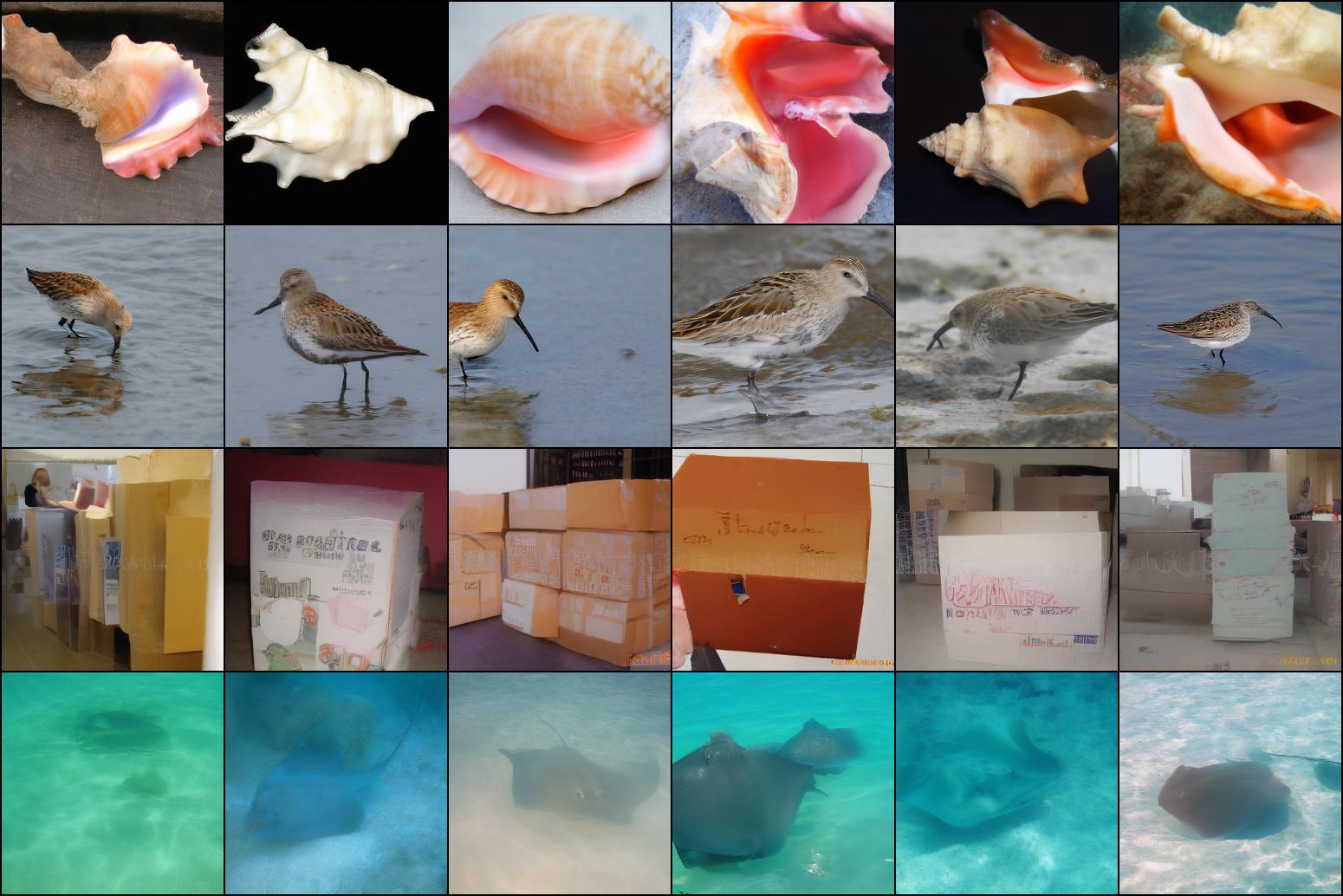}
        \subcaption{PTQD (w4a8)}
    \end{subfigure}
     \begin{subfigure}[tp!]{0.55\textwidth}
        \centering
        \includegraphics[width=\textwidth]{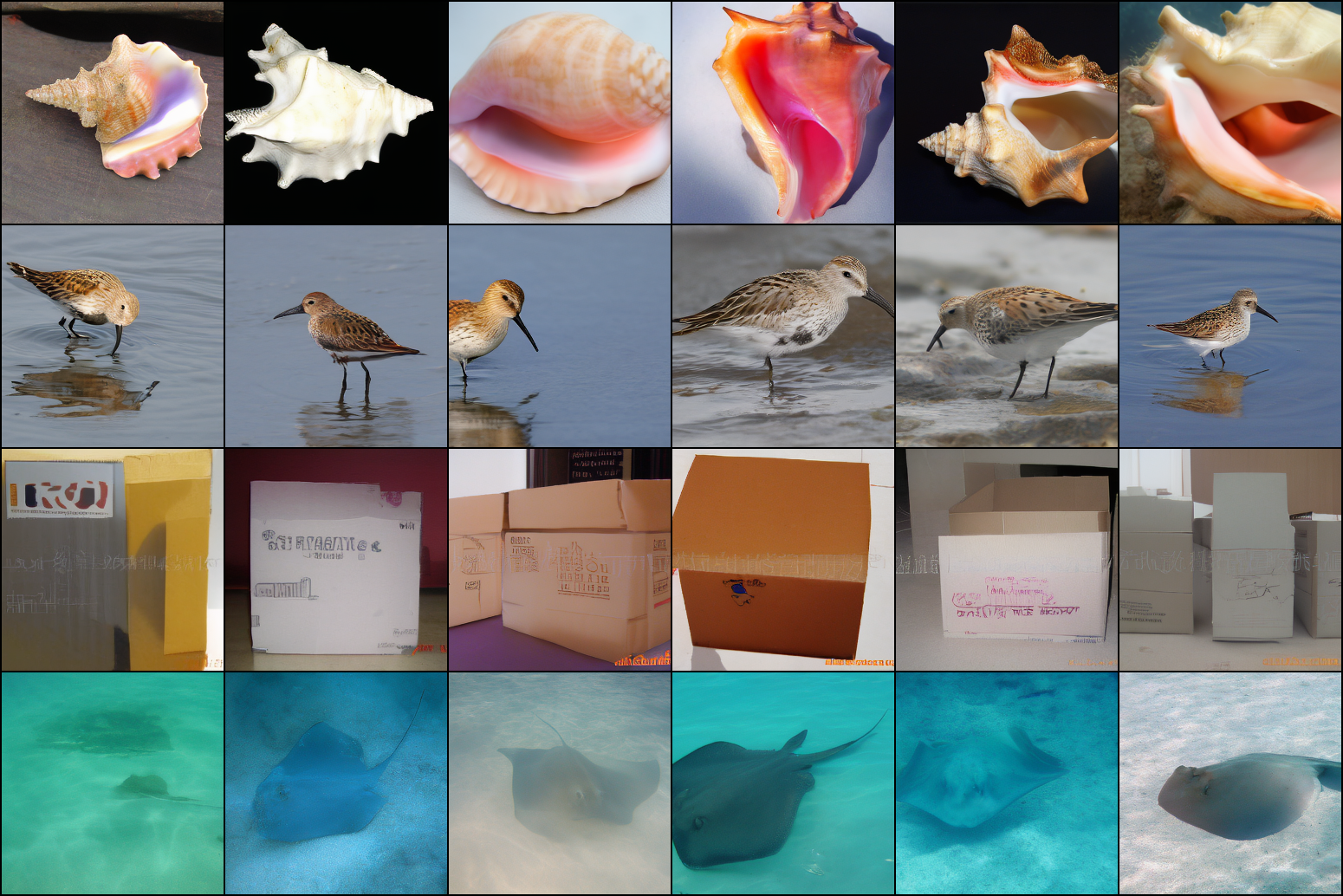}
        \subcaption{TFMQ-DM (w4a8)}
    \end{subfigure}
    \caption{Random samples from w4a8 quantized and full-precision LDM-4 on ImageNet $256\times 256$. The resolution of each sample is $256\times256$.}
\end{figure*}
\begin{figure*}[tp!]
   \centering
    \setlength{\abovecaptionskip}{0.2cm}
    \begin{subfigure}[tp!]{0.5\textwidth}
        \centering
        \includegraphics[width=\textwidth]{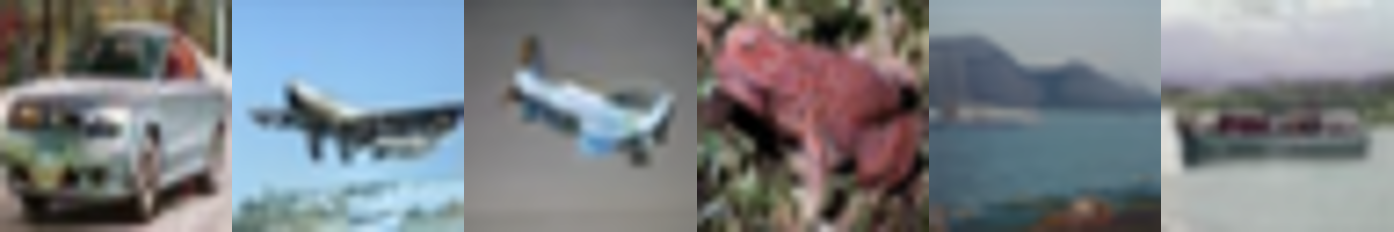}
        \subcaption{FP}
    \end{subfigure}
    \begin{subfigure}[tp!]{0.5\textwidth}
        \centering
        \includegraphics[width=\textwidth]{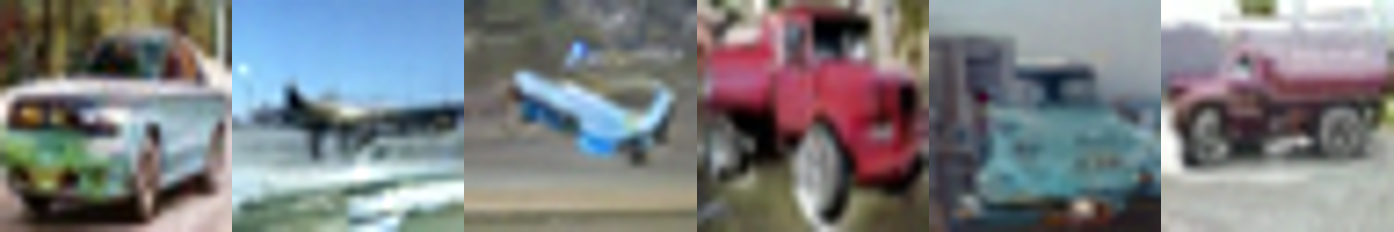}
        \subcaption{Q-Diffusion (w4a8)}
    \end{subfigure}
    \begin{subfigure}[tp!]{0.5\textwidth}
        \centering
        \includegraphics[width=\textwidth]{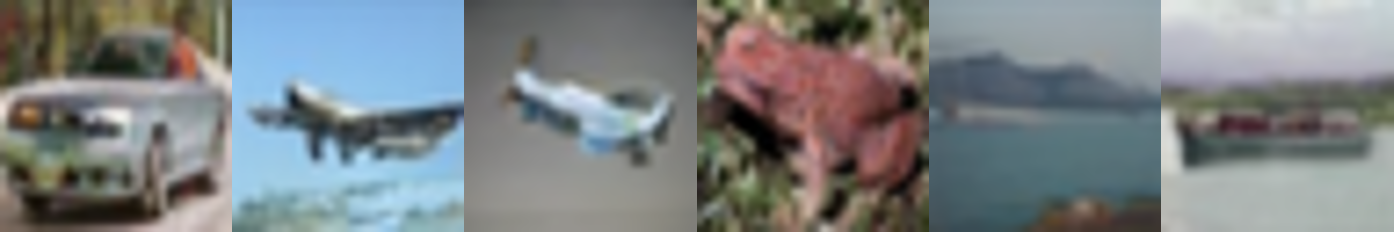}
        \subcaption{TFMQ-DM (w4a8)}
    \end{subfigure}
    \caption{Random samples from w4a8 quantized and full-precision DDIM on CIFAR-10 $32 \times 32$. The resolution of each sample is $32\times 32$.}
\end{figure*}

\end{document}